\documentclass{article} 
\usepackage{iclr2027_conference,times}

\usepackage{amsmath,amsfonts,bm}

\def\eqref#1{equation~\ref{#1}}

\def\1{\bm{1}}

\DeclareMathAlphabet{\mathsfit}{\encodingdefault}{\sfdefault}{m}{sl}
\SetMathAlphabet{\mathsfit}{bold}{\encodingdefault}{\sfdefault}{bx}{n}

\usepackage{hyperref}
\usepackage{url}
\usepackage{graphicx}
\usepackage{booktabs} 
\usepackage{wrapfig}
\usepackage{caption}
\usepackage{rotating}
\usepackage{lscape}
\usepackage{longtable}
\usepackage{listings}
\usepackage{graphicx} 
\definecolor{promptink}{HTML}{172636}
\definecolor{promptline}{HTML}{8FA6BC}
\definecolor{promptwash}{HTML}{F4F7FA}
\definecolor{prompttag}{HTML}{315C7D}
\newcommand{\prompttitle}[1]{%
  \begingroup
  \setlength{\fboxsep}{3.2pt}%
  \colorbox{prompttag}{\textcolor{white}{\sffamily\bfseries\scriptsize PROMPT}}%
  \hspace{0.55em}{\sffamily\bfseries\small\textcolor{promptink}{#1}}%
  \endgroup}
\lstdefinestyle{physcountprompt}{
  basicstyle=\ttfamily\fontsize{7.5}{8.9}\selectfont\color{promptink},
  backgroundcolor=\color{promptwash},
  rulecolor=\color{promptline},
  frame=single,
  framerule=0.55pt,
  frameround=tttt,
  framesep=7pt,
  breaklines=true,
  breakatwhitespace=false,
  columns=fullflexible,
  keepspaces=true,
  xleftmargin=0.25em,
  xrightmargin=0.25em,
  framexleftmargin=0.35em,
  framexrightmargin=0.35em,
  aboveskip=0.9em,
  belowskip=0.85em,
  showstringspaces=false
}
\lstnewenvironment{PromptBlock}[1]{%
  \lstset{style=physcountprompt,title={\prompttitle{#1}}}%
}{}
\definecolor{traceink}{HTML}{27313A}
\definecolor{tracewash}{HTML}{FBFCFD}
\definecolor{tracerule}{HTML}{C4CED7}
\definecolor{responsewash}{HTML}{F1F6F8}
\definecolor{tracecorrect}{HTML}{26705C}
\definecolor{traceincorrect}{HTML}{A14D45}
\newcommand{\tracebadge}[2]{%
  \begingroup\setlength{\fboxsep}{2.3pt}%
  \colorbox{#1!11}{\textcolor{#1}{\sffamily\bfseries\scriptsize #2}}%
  \endgroup}
\newcommand{\traceentryheader}[4]{%
  \par\vspace{0.55em}\noindent
  \begin{minipage}{\linewidth}
  \sffamily\small
  \textbf{#1}\hfill
  \textcolor{traceink!78}{Prediction: \textbf{#2}}\hspace{0.65em}%
  \tracebadge{#3}{#4}
  \end{minipage}\par\vspace{0.18em}}
\newcommand{\tracelistingtitle}[1]{%
  \sffamily\bfseries\scriptsize\textcolor{prompttag}{#1}}
\lstdefinestyle{physcounttrace}{
  basicstyle=\ttfamily\fontsize{7.65}{9.45}\selectfont\color{traceink},
  backgroundcolor=\color{tracewash},
  rulecolor=\color{tracerule},
  frame=single,
  framerule=0.42pt,
  frameround=tttt,
  framesep=7pt,
  breaklines=true,
  breakatwhitespace=false,
  columns=fixed,
  basewidth=0.52em,
  keepspaces=true,
  breakindent=0pt,
  breakautoindent=false,
  resetmargins=true,
  xleftmargin=0pt,
  framexleftmargin=6pt,
  xrightmargin=0pt,
  framexrightmargin=6pt,
  aboveskip=0.15em,
  belowskip=0.75em,
  showstringspaces=false,
  literate={—}{{---}}1 {–}{{--}}1 {’}{{'}}1 {×}{{$\times$}}1 {≈}{{$\approx$}}1 {°}{{$^\circ$}}1
}
\lstdefinestyle{physcountresponse}{
  basicstyle=\ttfamily\bfseries\fontsize{8.1}{9.6}\selectfont\color{promptink},
  backgroundcolor=\color{responsewash},
  rulecolor=\color{promptline},
  frame=single,
  framerule=0.42pt,
  frameround=tttt,
  framesep=6pt,
  breaklines=true,
  columns=fixed,
  basewidth=0.52em,
  keepspaces=true,
  breakindent=0pt,
  breakautoindent=false,
  resetmargins=true,
  xleftmargin=0pt,
  framexleftmargin=6pt,
  xrightmargin=0pt,
  framexrightmargin=6pt,
  aboveskip=0.1em,
  belowskip=1.0em,
  showstringspaces=false,
  literate={—}{{---}}1 {–}{{--}}1 {’}{{'}}1 {×}{{$\times$}}1 {≈}{{$\approx$}}1 {°}{{$^\circ$}}1
}
\definecolor{physdirect}{HTML}{0072B2}
\definecolor{physreflect}{HTML}{D55E00}
\definecolor{phystransmit}{HTML}{009E73}
\newcommand{\modechip}[2]{%
  \begingroup\setlength{\fboxsep}{2.2pt}%
  \colorbox{#1!12}{\textcolor{#1}{\bfseries #2}}%
  \endgroup}
\title{PhysCount: Do MLLMs Count Visual Appearances or Physical Entities?}

\author{Lidong Lu$^{1}$, Jiaxin Lv$^{1}$, Liangrui Dong$^{1}$, Xinyi Mao$^{1}$,
Lidong Zou$^{1}$, Zhenquan Li$^{1}$, \\ \textbf{Tong Lu$^{1}$, Guo Chen$^{2}$} \\[3pt]
\normalfont $^{1}$Nanjing University \qquad $^{2}$NVIDIA \\[4pt]
\makebox[\dimexpr\textwidth-2\tabcolsep\relax][c]{\small \url{https://PhysCount.github.io}}}

\iclrfinalcopy
\begin{document}

\maketitle
\fancyhead{}
\renewcommand{\headrulewidth}{0pt}

\begin{abstract}
Images reveal appearances, but counting often asks about physical objects. A single entity may be visible directly, in several reflections, or through glass, so detecting every visible occurrence does not determine how many entities are present. Existing visual counting benchmarks rarely separate these two operations. We introduce PhysCount, a benchmark for evaluating whether multimodal large language models (MLLMs) can distinguish what appears in an image from what physically exists in the scene. PhysCount contains 546 real-world images and 4,595 human-written questions organized into 651 target-specific groups. For each group, annotators locate the relevant appearances, label how each is observed, and identify appearances of the same entity. Four complementary question families test the enumeration of appearances, the enumeration of physical entities, image-formation-specific counting, and appearance--entity correspondence. We measure both accuracy on each family and the proportion of groups answered entirely correctly. Across 40 MLLMs, counting visible appearances is substantially easier than recovering the identities that connect them. Even the strongest model answers every question correctly in only 13.36\% of groups. Performance declines when one entity produces many appearances or when its appearances cross observation modes. Higher image resolution improves access to visual evidence but brings limited gains in complete-group accuracy, whereas answering related questions jointly produces a marked improvement. These results expose a gap between recognizing what is visible and correctly recovering the physical entities behind those observations.

\end{abstract}

\section{Introduction}

Intelligent perception requires recovering a stable physical world from changing and sometimes misleading appearances. Developmental psychology studies this ability through the appearance--reality distinction: children gradually learn that changes in how an object looks do not imply changes in what the object is~\cite{Flavell_1983}. Theories of object perception describe a similar process in which observations are organized around persistent physical units~\cite{Spelke_1990}, with ``object files'' connecting different appearances that are understood as states of the same object~\cite{Kahneman_1992}. The importance of this ability becomes especially clear when visible evidence conflicts with physical structure. Mirror self-recognition experiments test whether an animal relates a reflected body to itself~\cite{Gallup_1970}, while a transparent barrier leaves a target fully visible but physically inaccessible~\cite{Kabadayi_2017}. In both cases, successful behavior depends on interpreting the observation through the entities and relations that produce it. Sensory appearances provide evidence about the world, but they are not themselves the physical units into which the world is organized.

This distinction is equally consequential for multimodal large language models (MLLMs). A mirror can make one object appear multiple times in an image without creating additional entities, and an object seen through glass remains a physical entity despite the intervening surface. A model may therefore identify every visible instance and still answer incorrectly if it treats each appearance as a separate thing. Counting in such scenes requires more than locating targets and adding them up; the model must determine which observations arise from the same entity. This leads to the central question of this work: \emph{Do MLLMs count visual appearances or physical entities?}

\begin{figure}[h]
    \hyphenpenalty=10000\exhyphenpenalty=10000
    \centering
    \includegraphics[width=\linewidth]{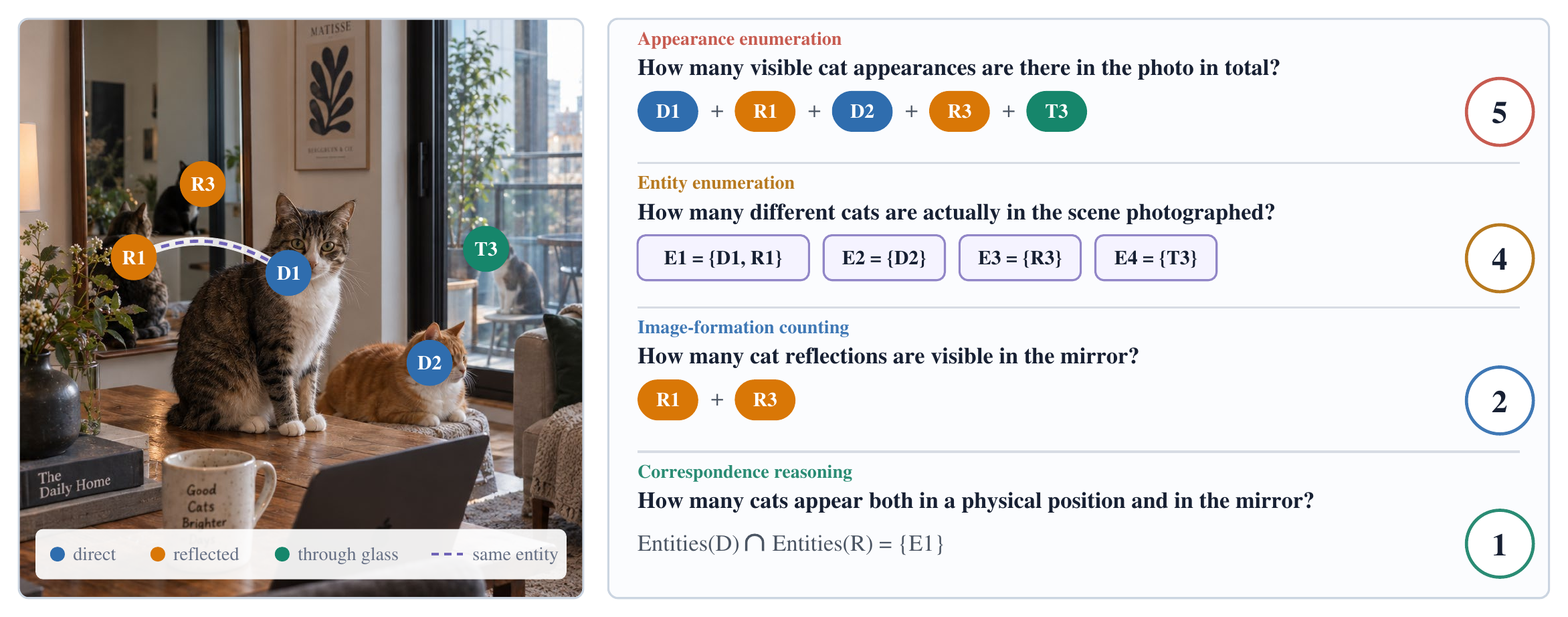}
    \caption{\textbf{One scene, different counting units.}
    Five cat appearances correspond to four entities because D1 and R1 depict the same cat. With the image and target fixed, the questions count all appearances (5), distinct entities (4), reflections (2), or entities seen both directly and in reflection (1).}
    \label{fig:teaser}
\end{figure}

Existing visual counting tasks typically begin with a target category and treat visible instances as the units to be counted, whether through question answering, category-agnostic counting, language-guided density estimation, or multimodal numerosity evaluation~\cite{Acharya_2019,Ranjan_2021_CVPR,Jiang_2023,Paiss_2023_ICCV}. These formulations evaluate whether a model can locate and enumerate the requested targets, while leaving the counting unit implicit. Mirrors, transparent surfaces, and repeated imaging paths separate visible appearances from physical entities. One entity may produce several spatially distinct observations, so the same visual inventory can support different counts depending on how those observations are grouped. Answering ``how many'' therefore requires both recovering the appearances and determining which of them share a physical source.

We introduce PhysCount, a benchmark for testing whether MLLMs can distinguish visible appearances from the physical entities they depict. Built from 546 natural images, PhysCount comprises 651 target-specific groups and 4,595 questions. For each counting target, annotators mark all relevant appearances, record whether each is seen directly, in reflection, or through glass, and group appearances that depict the same entity. These annotations generate four types of questions about appearances, physical entities, image formation, and correspondence. Questions associated with the same image and target probe complementary quantities derived from one annotated scene. PhysCount evaluates each count separately and reports how often a model solves the complete question group.

Experiments with 40 MLLMs reveal a clear separation between appearance counting and questions that depend on physical identity. Errors increase when the correct count requires grouping repeated appearances or matching observations across different modes. Scaling and higher resolution improve individual answers more readily than they improve complete-group accuracy. Presenting related questions together substantially improves both aggregate accuracy and the number of groups solved in full. Across these results, the largest remaining difficulty lies in counts defined by appearance--entity correspondence rather than in appearance enumeration alone.

Our contributions can be summarized as follows:
\begin{itemize}
\item We identify an overlooked distinction in visual counting: visible appearances and physical entities are not always in one-to-one correspondence, so a correct numerical answer does not necessarily show that a model has counted the intended units.

\item We introduce PhysCount, comprising 546 natural images, 651 target-specific groups, and 4,595 questions. Its annotations record appearance locations, image-formation modes, and same-entity assignments, supporting questions about appearances, physical entities, image formation, and correspondence, together with strict complete-group evaluation.

\item We evaluate 40 representative MLLMs and show that appearance counts are more reliable than counts requiring physical-identity correspondence. Performance declines on repeated and cross-mode appearances, while joint question presentation improves complete-group accuracy.
\end{itemize}

\section{Related Work}
\paragraph{Visual Counting.}
Visual counting benchmarks have broadened both the form of supervision and the range of scenes under evaluation. TallyQA frames counting as question answering in natural images, while category-agnostic and language-guided methods study transfer to previously unseen targets~\cite{Acharya_2019,Ranjan_2021_CVPR,Jiang_2023,Paiss_2023_ICCV}. Recent multimodal benchmarks further examine dense and open-world counting, numerical question answering, and the counting behavior of general-purpose vision--language models~\cite{Mondal_2025,Deitke_2025_CVPR,tamarapalli2025countqamllmscountwild,Weng_2025_ICCV,deng2026holocountholisticvisualcounting}. Other work changes the information that must be selected or inferred: MIND and PairTally emphasize linguistic target specification~\cite{Ciampi_2025_WACV,nguyen2025currentaimodelscount}, CAPTURe recovers counts under occlusion~\cite{Pothiraj_2025_ICCV}, and video or audio-visual benchmarks extend enumeration across time and modalities~\cite{Lu_2026_CVPR,tsuchiya2026diagnosinglongvideoquantitativereasoning,Rong_2026_CVPR}. This progression has made counting more open-ended, compositional, and robust to incomplete evidence, but the unit being counted is still generally a visible category instance or an instance inferred to occupy the scene. Work on numerical cognition offers a complementary view: small-set judgments and approximate numerosity are distinct processes, while object-based accounts tie number to representations that persist across changes in view~\cite{Mandler_1982,Feigenson_2004,Wynn_1992,Kahneman_1992}. PhysCount extends this distinction to MLLMs by evaluating two abilities separately: enumerating visible appearances and grouping appearances that depict the same physical entity.

\paragraph{Mirror Understanding in MLLMs.}
Recent benchmarks examine mirror understanding in MLLMs from two distinct perspectives. MirrorQA evaluates subject-centered left--right reasoning in mirror images, testing whether models correctly transform spatial orientation~\cite{liu-etal-2026-mirrorqa}. MirrorBench adopts an embodied perspective: inspired by the mirror self-recognition paradigm, it evaluates self-centric intelligence from basic mirror perception to higher-level self-representation in simulated environments~\cite{guo2026mirrorbenchevaluatingselfcentricintelligence}. PhysCount targets a different capability. Rather than asking models to infer orientation or recognize themselves, it asks them to enumerate the target appearances in a scene and determine which of those appearances depict the same physical entity. This requires handling mirror-only entities, direct views paired with reflections, and multiple reflections generated by one entity. PhysCount represents these cases explicitly by annotating each appearance, its image-formation mode, and its physical-identity group, then evaluates the resulting appearance, entity, and correspondence counts.

\section{PhysCount: Counting Beyond Visible Appearances}
\subsection{Task Formulation}
PhysCount presents a model with an image and a target-specific counting question and expects an integer answer. The number of visible target appearances may differ from the number of objects in the scene. The same object may be seen directly and in one or more mirrors, while another may be visible only through reflection or glass. The question determines whether the model should count each visible appearance or merge appearances that depict the same physical entity.

PhysCount evaluates four question families: counting all visible appearances, counting physical entities, counting a specified type of appearance, and reasoning about appearance--entity correspondences. The same image can consequently admit several correct counts depending on what the question specifies. Together, these questions separate the recovery of visual evidence from the interpretation of that evidence in terms of physical entities.

\subsection{Data Collection and Annotation}
\label{sec:data_collection}
\begin{figure}[h]
    \hyphenpenalty=10000\exhyphenpenalty=10000
    \centering
    \includegraphics[width=\linewidth]{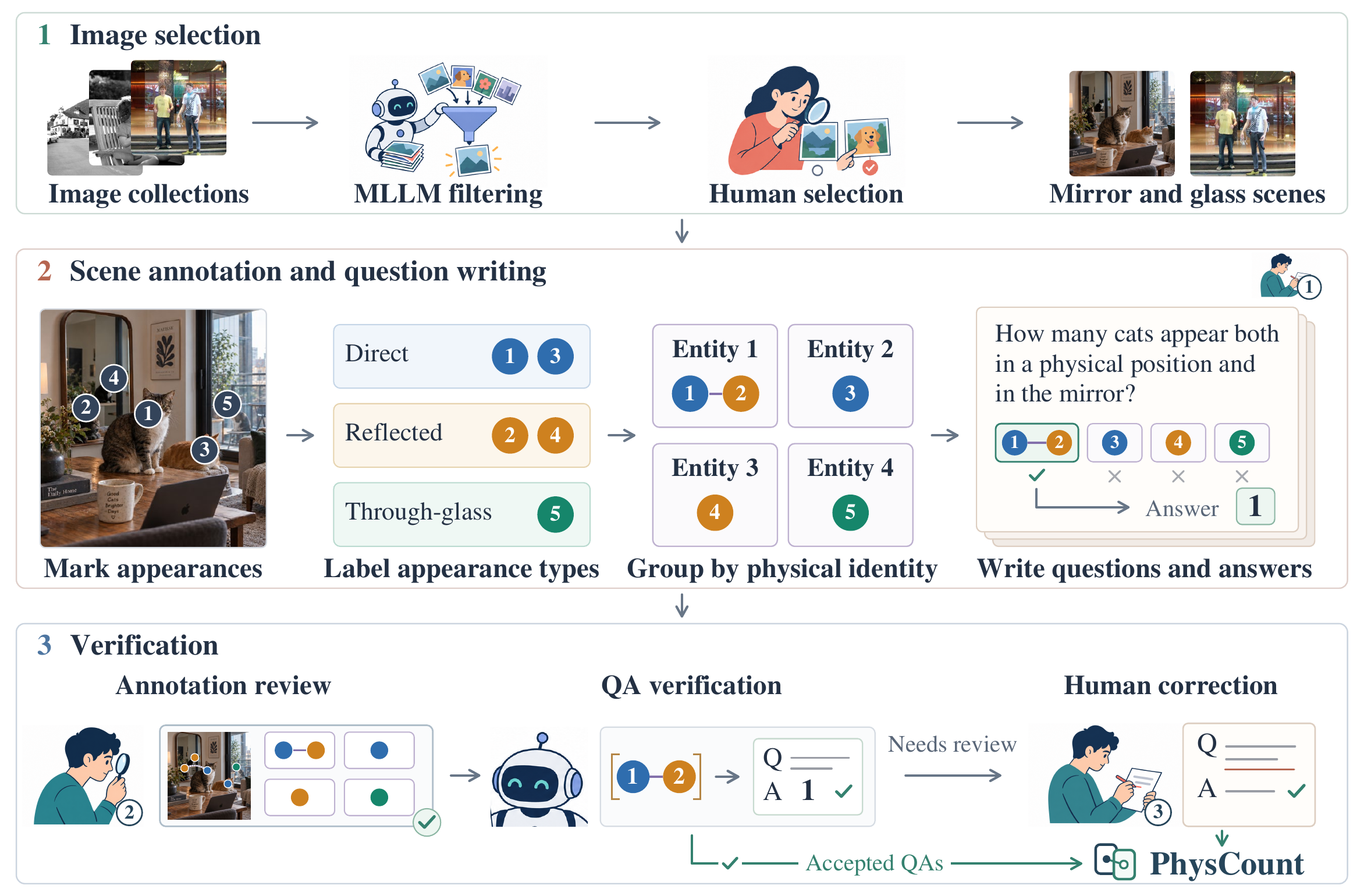}
    \caption{\textbf{PhysCount construction pipeline.} Candidate mirror and glass scenes undergo automatic filtering and human selection. An annotator labels direct, reflected, and through-glass appearances, groups appearances of the same entity, and writes questions whose answers are derived from these groups. Here, five cat appearances form four entities, and the correspondence question selects the one entity seen both directly and in reflection. Annotation review and automated QA checking send consistent QAs to PhysCount and flagged QAs to final human correction.}
    \label{fig:data_pipeline}
\end{figure}

Figure~\ref{fig:data_pipeline} summarizes the construction of PhysCount in three stages. First, we build the image set through automatic candidate filtering followed by human selection. Qwen3.6-35B-A3B~\cite{qwen36_35b_a3b} filters natural images from Common Canvas~\cite{Gokaslan_2024_CVPR}, GPIC~\cite{chandrasegaran2026gpicgiantpermissiveimage}, and other public collections for mirror and glass scenes in which reflections or glass change the apparent count. Human annotators then inspect these candidates and select images with localizable targets, distinguishable direct, reflected, or through-glass appearances, and counts that can be resolved through careful inspection. The selected images contain salient appearance duplication or interactions between reflected and transmitted content.

Second, annotators construct a target-specific scene representation. An image containing several suitable target categories produces a separate annotation group for each category. Within a group, annotators mark every visible target occurrence and label it as direct, reflected, or seen through glass. Occurrences depicting the same real-world object receive a common identity. Thus, each point records the location and type of one appearance, while each identity group records the physical entity depicted by one or more appearances. The grouping covers direct--reflection pairs, objects visible only in mirrors, and multiple reflections of one object.

Finally, the first annotator writes counting questions and their integer answers based on the completed scene annotation. A second annotator checks the appearance locations, appearance-type labels, and identity groups against the image. GPT-5.6 Sol then uses the verified annotation to check whether each human-written question can be answered from the recorded information and whether its answer gives the required count. A third annotator reviews and corrects every flagged QA pair. Appendix~\ref{app:annotation} provides the complete instructions and examples.

\subsection{Question Design}
PhysCount uses four question families to expose different failure points between seeing a target and recovering the physical scene. The image and target remain fixed while the required count changes, so success depends on selecting the correct counting unit and applying the required operation to the same annotated evidence. Figure~\ref{fig:teaser} shows how the same scene yields different answers as the requested counting unit changes.

\emph{Appearance} questions ask for every visible occurrence of the target, including direct, reflected, and through-glass appearances. The difficulty is to identify separate occurrences before deciding whether they share an identity. Reflections may be cropped, distorted, or repeated across mirrors, while reflected and transmitted content may overlap on the same glass surface. A model that merges similar-looking regions too early will miss valid appearances; one that splits a single occurrence into fragments will overcount them. Each annotation group includes one appearance question to measure this initial visual inventory.

\emph{Entity} questions ask how many physical objects are represented by that inventory. This requires matching appearances despite changes in position, scale, orientation, visibility, and appearance type. A direct observation and several mirror images may all depict one object, whereas two similar reflections may belong to different objects. Mirror-only entities add a further challenge because identity must be established without a direct appearance as an anchor. The resulting errors have two opposite forms: counting one object several times or merging distinct objects into one.

\emph{Image-formation} questions ask whether each appearance is seen directly, formed by reflection, or observed through glass. Object recognition alone does not determine this label: the same person or object can be visible directly, in reflection, and through glass in one image. Glass is especially demanding because reflected foreground content and transmitted background content occupy the same pixels, and their boundaries may cross or partly hide one another. The model must separate these layers and determine how each target occurrence becomes visible.

\emph{Correspondence} questions examine whether the recovered appearances, appearance types, and identities agree. They count intersections, exclusive sets, and multiplicities, such as entities seen both directly and in reflection or entities producing several mirror images. These answers depend on the full pattern of identity assignments: changing one appearance--entity match can alter several relation counts at once. The challenge is therefore not a single local match, but maintaining one consistent set of identities across all observed appearances.

\subsection{Benchmark Statistics}

\begin{figure}[h]
\hyphenpenalty=10000\exhyphenpenalty=10000
\captionsetup{font=footnotesize,skip=3pt}
\centering
\includegraphics[width=\linewidth]{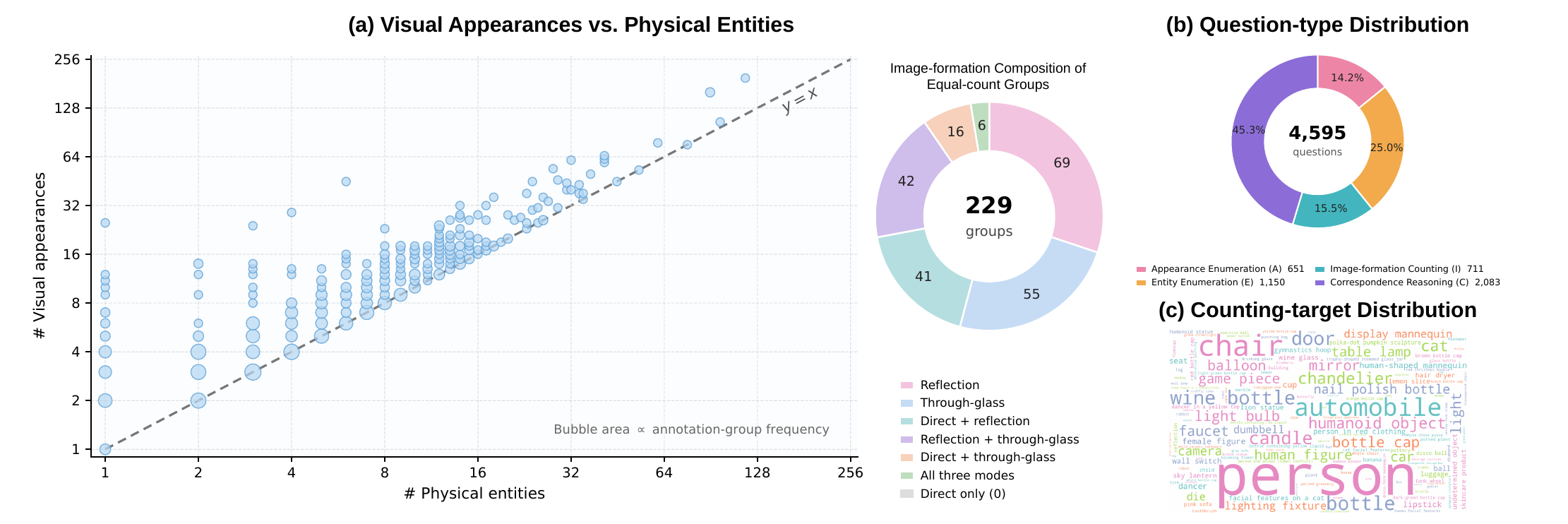}
\caption{\textbf{Benchmark statistics.} (a) Visual-appearance and physical-entity counts across 651 annotation groups; bubble area encodes the number of groups at each coordinate, and the adjacent chart shows the appearance-type composition of the 229 groups on $y=x$. (b) Distribution of 4,595 questions across the four question families. (c) Diversity and frequency of counting targets.}
\label{fig:benchmark_statistics}
\end{figure}

PhysCount contains 546 images, 651 target-specific annotation groups, and 4,595 factual questions covering 137 normalized counting targets. Every group includes an appearance question, and 599 groups include all four question families described above. A group contains 7.06 questions on average. As shown in Figure~\ref{fig:benchmark_statistics}(b), correspondence is the largest family, with 2,083 questions: 924 concern visibility sets, 459 ask about overlap among direct, reflected, and through-glass appearances, and 700 examine how many appearances are associated with each entity.

The annotations contain 7,086 target appearances, including 2,416 direct appearances, 3,996 reflections, and 674 appearances seen through glass. Among the 651 groups, 475 include an entity visible through reflection or glass with no direct appearance. Figure~\ref{fig:benchmark_statistics}(a) relates the number of annotated appearances to the number of physical entities. The 229 groups on $y=x$ contain one appearance per entity; 166 contain indirect appearances alone, while 63 combine direct and indirect appearances belonging to different entities. The other 422 groups contain at least one entity with multiple appearances. In 402 groups, the same entity has appearances of different types, such as a direct appearance and a reflection. In the other 20, the same entity appears several times, but all of its appearances have the same type, as when one object produces several reflections.

Figure~\ref{fig:benchmark_statistics}(c) shows the diversity and long tail of the 137 counting targets. They include people, animals, vehicles, furniture, containers, lighting, apparel, food, plants, toys, architectural elements, and many less frequent everyday objects. Answer values range from 0 to 196 and differ across question families: appearance questions have a mean answer of 10.88, while correspondence questions usually count smaller sets defined by identity relations.

\begin{table}[t]
\caption{\textbf{Results on PhysCount for all 40 evaluated models and five human participants.} Each participant independently answers all 4,595 questions under the same image--question protocol as the models; the human row reports their mean. Best and second-best model results are bolded and underlined.}
\label{tab:main_results}
\centering
\scriptsize
\renewcommand{\arraystretch}{1.08}
\setlength{\tabcolsep}{2.35pt}
\resizebox{\textwidth}{!}{%
\begin{tabular}{lrrrrrrrrrrr}
\toprule
& \multicolumn{2}{c}{Appearance}
& \multicolumn{2}{c}{Entity}
& \multicolumn{2}{c}{Image Formation}
& \multicolumn{2}{c}{Correspondence}
& & & \\
\cmidrule(lr){2-3}\cmidrule(lr){4-5}\cmidrule(lr){6-7}\cmidrule(lr){8-9}
Model
& MAE$\downarrow$ & NAE$\downarrow$
& MAE$\downarrow$ & NAE$\downarrow$
& MAE$\downarrow$ & NAE$\downarrow$
& MAE$\downarrow$ & NAE$\downarrow$
& Score$\uparrow$ & Group-Acc$\uparrow$ & Phys-Gnd$\uparrow$ \\
\midrule
\textit{Human (5-person avg.)} & 0.037 & 0.59 & 0.023 & 0.83 & 0.035 & 1.05 & 0.033 & 2.21 & 97.10 & 95.98 & 99.49 \\
\midrule
Gemini 3.7 Flash~\cite{gemini37flash} & \textbf{2.724} & 20.08 & \textbf{2.046} & \textbf{27.36} & 2.467 & 30.62 & \textbf{1.396} & \textbf{33.88} & \textbf{36.45} & \textbf{13.36} & \textbf{73.00} \\
GPT-5.6 Sol~\cite{openai2026gpt56} & 3.028 & 22.82 & 2.808 & 36.58 & 2.544 & 31.94 & 2.113 & 52.11 & 21.73 & 3.84 & 55.65 \\
GPT-6 Astra~\cite{openai2026gpt6astra} & 3.049 & 24.87 & 2.673 & 37.72 & 2.517 & 30.99 & 1.905 & 47.83 & 20.77 & 3.38 & 51.89 \\
GPT-5.6 Terra~\cite{openai2026gpt56} & 3.395 & 26.06 & 2.726 & 36.39 & 3.227 & 37.30 & 1.863 & 50.99 & 20.42 & 4.30 & 53.43 \\
GPT-5.6 Luna~\cite{openai2026gpt56} & 3.536 & 28.24 & 3.368 & 42.24 & 2.992 & 35.07 & 2.283 & 55.98 & 16.87 & 2.92 & 46.98 \\
\midrule
Qwen3.8-27B~\cite{qwen38} & 3.149 & 19.89 & 2.511 & 32.42 & 2.702 & 29.66 & 1.685 & 44.68 & 28.84 & 7.37 & \underline{65.96} \\
Qwen3.5-397B-A17B~\cite{qwen3.5} & \underline{2.836} & 18.99 & 2.442 & 33.28 & \textbf{2.386} & \underline{26.88} & 1.859 & 46.73 & 27.95 & 5.84 & 62.16 \\
Qwen3.5-122B-A10B~\cite{qwen3.5} & 4.192 & \textbf{17.96} & 2.527 & 34.40 & 2.603 & 27.80 & 2.068 & 49.00 & 26.03 & 5.99 & 57.13 \\
Qwen3.5-35B-A3B~\cite{qwen3.5} & 3.143 & \underline{18.81} & \underline{2.425} & 33.44 & 2.621 & 28.49 & 2.011 & 51.51 & 25.70 & 5.99 & 58.76 \\
Qwen3.5-27B~\cite{qwen3.5} & 4.103 & 20.64 & 3.216 & 34.58 & 3.017 & \textbf{26.78} & 2.942 & 48.56 & 26.32 & 6.30 & 60.83 \\
Qwen3.5-9B~\cite{qwen3.5} & 2.845 & 20.04 & 3.708 & 36.09 & 3.805 & 30.05 & 4.816 & 53.38 & 23.54 & 5.07 & 55.41 \\
Qwen3.5-4B~\cite{qwen3.5} & 3.321 & 22.19 & 4.379 & 39.57 & 2.729 & 32.93 & 3.274 & 57.03 & 18.57 & 3.38 & 48.20 \\
\midrule
Gemma 4-31B~\cite{gemmateam2026gemma4} & 3.307 & 21.83 & 2.700 & 36.88 & 2.814 & 33.71 & 1.879 & \underline{41.86} & 26.32 & 6.61 & 60.07 \\
Gemma 4-26B-A4B~\cite{gemmateam2026gemma4} & 3.257 & 22.70 & 2.852 & 39.22 & 2.936 & 36.53 & 1.888 & 44.02 & 23.79 & 5.53 & 57.17 \\
Gemma 4-12B~\cite{gemmateam2026gemma4} & 3.490 & 26.05 & 3.044 & 44.56 & 3.057 & 37.08 & 1.982 & 45.70 & 19.03 & 3.07 & 46.10 \\
Gemma 4-E4B~\cite{gemmateam2026gemma4} & 37.605 & 50.27 & 22.506 & 64.37 & 22.469 & 69.95 & 9.344 & 65.25 & 2.93 & 0.00 & 11.90 \\
Gemma 4-E2B~\cite{gemmateam2026gemma4} & 20.343 & 48.02 & 8.357 & 55.57 & 11.088 & 55.82 & 4.146 & 62.93 & 4.76 & 0.15 & 21.13 \\
\midrule
InternVL3.5-241B-A28B~\cite{wang2025internvl35advancingopensourcemultimodal} & 4.135 & 30.21 & 2.825 & 36.24 & 4.597 & 36.63 & 1.983 & 48.11 & 21.04 & 4.92 & 60.27 \\
InternVL3.5-38B~\cite{wang2025internvl35advancingopensourcemultimodal} & 4.078 & 32.36 & 4.520 & 39.75 & 10.133 & 36.59 & 3.837 & 57.92 & 15.55 & 2.61 & 47.23 \\
InternVL3.5-30B-A3B~\cite{wang2025internvl35advancingopensourcemultimodal} & 4.152 & 29.51 & 3.036 & 38.94 & 3.359 & 37.29 & 2.108 & 52.08 & 17.76 & 2.61 & 53.91 \\
InternVL3.5-14B~\cite{wang2025internvl35advancingopensourcemultimodal} & 4.158 & 29.13 & 2.855 & 40.02 & 3.483 & 38.22 & 2.099 & 57.32 & 15.20 & 1.84 & 44.22 \\
InternVL3.5-8B~\cite{wang2025internvl35advancingopensourcemultimodal} & 3.938 & 27.82 & 3.011 & 41.68 & 3.670 & 42.16 & 2.233 & 57.78 & 14.29 & 1.84 & 47.08 \\
InternVL3.5-4B~\cite{wang2025internvl35advancingopensourcemultimodal} & 7.018 & 38.11 & 3.452 & 48.83 & 3.685 & 39.85 & 2.773 & 65.97 & 9.53 & 1.23 & 36.93 \\
\midrule
Muse-Glimmer-30B~\cite{meta2026museglimmer} & 2.943 & 20.31 & 2.478 & \underline{32.10} & \underline{2.405} & 28.69 & \underline{1.608} & 43.04 & \underline{29.61} & \underline{7.83} & 65.51 \\
DeepSeek-V4-Flash-Vision~\cite{deepseek2026v4vision} & 4.321 & 30.52 & 2.814 & 36.12 & 3.406 & 37.19 & 1.864 & 48.56 & 20.74 & 5.22 & 61.61 \\
MiniMax-M3-428B-A23B~\cite{lai2026minimaxsparseattention} & 3.888 & 29.28 & 2.989 & 39.05 & 3.739 & 39.57 & 1.975 & 52.65 & 17.43 & 3.69 & 52.50 \\
MiniCPM-V 4.5-8B~\cite{yu2025minicpmv45cookingefficient} & 4.528 & 32.52 & 2.828 & 38.82 & 3.426 & 39.32 & 2.176 & 58.40 & 14.90 & 2.76 & 51.28 \\
MiniCPM-V 4.6-1.3B~\cite{yao2024minicpm} & 5.720 & 39.90 & 3.340 & 43.56 & 3.694 & 42.36 & 2.252 & 57.22 & 12.40 & 2.15 & 45.11 \\
Cosmos3-Super-64B~\cite{nvidia2026cosmos3omnimodalworld} & 4.144 & 28.29 & 3.003 & 37.39 & 3.051 & 33.50 & 2.339 & 64.08 & 14.69 & 1.38 & 45.67 \\
Cosmos3-Nano-16B~\cite{nvidia2026cosmos3omnimodalworld} & 6.002 & 30.81 & 3.239 & 41.23 & 3.311 & 36.84 & 2.548 & 68.01 & 12.08 & 2.15 & 36.23 \\
Cosmos3-Edge-4B~\cite{nvidia2026cosmos3omnimodalworld} & 5.332 & 37.23 & 4.005 & 43.59 & 3.555 & 41.68 & 3.209 & 65.98 & 9.27 & 0.77 & 32.42 \\
Ovis2.6-80B-A3B~\cite{lu2024ovis} & 4.132 & 31.39 & 3.089 & 39.59 & 3.313 & 37.88 & 2.302 & 58.70 & 14.93 & 2.15 & 46.79 \\
Ovis2.6-30B-A3B~\cite{lu2024ovis} & 8.207 & 58.71 & 5.285 & 66.79 & 5.397 & 64.98 & 2.404 & 63.92 & 3.80 & 0.46 & 32.38 \\
Molmo2-4B~\cite{Clark_2026_CVPR} & 4.284 & 27.98 & 3.442 & 38.45 & 3.372 & 34.68 & 2.200 & 63.45 & 14.96 & 1.84 & 47.21 \\
Molmo2-8B~\cite{Clark_2026_CVPR} & 4.174 & 29.30 & 3.209 & 37.50 & 2.996 & 34.35 & 2.484 & 65.41 & 13.13 & 0.92 & 42.99 \\
ERNIE 4.5 VL-28B-A3B~\cite{ernie2025technicalreport} & 4.682 & 32.62 & 3.269 & 43.29 & 3.621 & 40.53 & 2.537 & 58.93 & 13.08 & 1.69 & 42.92 \\
Phi-4-Reasoning-Vision-15B~\cite{aneja2026phi4reasoningvision15btechnicalreport} & 3.779 & 30.31 & 3.461 & 47.21 & 3.681 & 43.79 & 2.425 & 60.20 & 11.25 & 0.61 & 38.63 \\
LLaVA-OneVision 2-8B~\cite{an2026llavaonevision2nextgenerationperceptualintelligence} & 3.690 & 28.70 & 2.948 & 42.56 & 3.475 & 39.05 & 2.749 & 67.94 & 10.51 & 1.08 & 33.74 \\
Granite Vision 4.1-4B~\cite{granite-vision-4.1-4b} & 4.152 & 34.73 & 3.437 & 45.38 & 3.836 & 41.51 & 2.266 & 67.96 & 9.02 & 0.61 & 32.62 \\
Mistral Small 4-119B~\cite{mistral2026small4} & 5.582 & 44.59 & 3.526 & 53.29 & 3.908 & 49.46 & 2.660 & 69.42 & 5.87 & 0.00 & 26.18 \\
\bottomrule
\end{tabular}}
\end{table}

\subsection{Evaluation Protocol}
\label{sec:evaluation_protocol}
For a non-negative integer prediction $p$ and ground-truth count $y$, we report absolute error and bounded normalized absolute error,
\begin{equation}
\mathrm{AE}(p,y)=|p-y|,\qquad \mathrm{NAE}(p,y)=\min\!\left(1,\frac{|p-y|}{\max(y,1)}\right).
\end{equation}
Outputs that cannot be parsed as valid counts receive the maximum normalized error, while MAE is computed over parsed predictions. For each question family, we first average the NAE of that family's questions within each annotation group. We then average these group-level values over all groups containing that family. A group therefore contributes one value regardless of how many questions of that family it contains. Let $a_g^t$ denote one minus the within-group NAE for family $t$, and let $\mathcal T_g$ contain the families available in group $g$. The primary aggregate metric is
\begin{equation}
\mathrm{Score}=\frac{100}{|\mathcal G|}\sum_{g\in\mathcal G}\prod_{t\in\mathcal T_g}a_g^t.
\end{equation}
The product rewards reliable performance across all question families in a group: strong appearance enumeration cannot offset a failed entity or correspondence estimate. We additionally report \emph{Group-Acc}, the percentage of annotation groups for which every answer exactly matches its ground truth. It measures complete group-level correctness. \emph{Phys-Gnd} isolates physical reasoning after successful enumeration. Among groups covering all four question families, it retains those whose total-appearance answer is exact, computes the harmonic mean of entity, image-formation, and correspondence accuracy, and then averages across the retained groups. An eligible group receives zero if any of these three components is zero. Because the retained groups depend on which appearance questions a model answers correctly, Phys-Gnd measures conditional physical grounding rather than overall capability.

\section{Experiments}
\subsection{Evaluation Setting}
\label{sec:evaluation_setting}
We evaluate 40 proprietary and open-weight MLLMs under a common task protocol. Unless stated otherwise, each question is answered independently from the same image, preventing related questions in an annotation group from sharing previous answers. Each model uses the strongest reasoning or thinking setting available in its evaluated backend and returns a single integer. Five human participants independently complete all 4,595 questions using the same image--question inputs; their metrics are averaged to form the human baseline. We preserve the native image resolution up to a $4{,}096$-pixel long edge and proportionally downsample larger images. The resolution and joint-question studies retain the same model configuration and image processing. Appendix~\ref{app:evaluation_details} provides the prompts, backend-specific inference settings, and output-validation rules.

\subsection{Main Results}

Table~\ref{tab:main_results} shows that Gemini 3.7 Flash achieves the highest overall Score, while Muse-Glimmer-30B and Qwen3.8-27B are the strongest open models. No model leads every metric. Qwen3.5-397B-A17B has the lowest appearance error and performs strongly on image formation, whereas Gemma~4-31B is more competitive on correspondence. Neither advantage produces the best overall Score. In contrast, the five-participant human average reaches 97.10 Score and 95.98 Group-Acc, showing that careful human observers can resolve nearly all quantities in a group together. The results distinguish three related capabilities: detecting what is visible, recognizing how each appearance is formed, and recovering the physical entities behind those appearances. Strength in one does not ensure strength in the others.

Across models, enumerating visible appearances is consistently easier than answering entity and correspondence questions. These questions use the same scene and counting target, but the latter require deciding which direct, reflected, and through-glass appearances depict the same object. Phys-Gnd remains well below its ceiling even on groups with an exact appearance answer. A correct appearance total therefore does not reliably carry over to counts defined by physical identity. The benchmark identifies this separation at the level of answer behavior; integer outputs alone do not determine which internal computation produced an error.

Group-Acc sets a stricter requirement: every appearance, entity, image-formation, and correspondence question in a group must be answered exactly. Many models answer a substantial fraction of individual questions correctly, yet only Gemini solves more than 10\% of annotation groups in full. Stronger model variants generally reduce error within each question family, but their Group-Acc remains low. Increased model capacity therefore improves individual counts more readily than it produces a complete solution to all quantities derived from the same annotated scene.

\subsection{Detailed Analysis}
\label{sec:detailed_analysis}

\paragraph{Impact of Input Resolution.}
Table~\ref{tab:resolution_analysis} traces how additional visual detail affects the counting process. Increasing the long-edge resolution improves the overall Score from 21.04 to 28.84, but most gains occur before the highest resolution. Appearance and correspondence errors reach their lowest values at 2,048 pixels and change little thereafter, while Group-Acc stops improving after 1,024 pixels. The different saturation points separate visual access from complete task success. Higher resolution helps models recover visible appearances, whereas solving an entire group also requires exact answers to its entity, image-formation, and correspondence questions. Once the visible inventory stabilizes, further pixels contribute little to these identity-dependent counts.

\begin{minipage}{\linewidth}
\hyphenpenalty=10000\exhyphenpenalty=10000
\begin{minipage}[t]{0.55\textwidth}
\vspace{0pt}
\captionsetup{font=footnotesize,skip=3pt,position=top}
\captionof{table}{\textbf{Impact of input resolution:} Qwen3.8-27B under independent questioning at four maximum long-edge resolutions.}
\label{tab:resolution_analysis}
{\centering
\scriptsize
\renewcommand{\arraystretch}{0.91}
\setlength{\tabcolsep}{4.2pt}
\begin{tabular}{lrrrr}
\toprule
Metric & 512 & 1,024 & 2,048 & 4,096\\
\midrule
Appearance NAE$\downarrow$ & 28.36 & 21.53 & \textbf{19.42} & 19.89\\
Entity NAE$\downarrow$ & 37.95 & 33.88 & 33.09 & \textbf{32.42}\\
Formation NAE$\downarrow$ & 37.08 & 32.12 & 31.69 & \textbf{29.66}\\
Correspond. NAE$\downarrow$ & 49.31 & 46.85 & \textbf{44.22} & 44.68\\
Score$\uparrow$ & 21.04 & 26.41 & 27.86 & \textbf{28.84}\\
Group-Acc$\uparrow$ & 5.38 & \textbf{7.37} & \textbf{7.37} & \textbf{7.37}\\
Phys-Gnd$\uparrow$ & 62.50 & 63.10 & 61.26 & \textbf{65.96}\\
\bottomrule
\end{tabular}
\par}

\vspace{0.75em}
\captionof{table}{\textbf{Impact of question presentation:} Qwen3.8-27B at a 4,096-pixel long edge with questions presented independently or jointly by annotation group.}

\label{tab:joint_questioning}
{\centering
\scriptsize
\renewcommand{\arraystretch}{0.91}
\setlength{\tabcolsep}{10pt}
\begin{tabular}{lrr}
\toprule
Metric & Independent & Joint\\
\midrule
Appearance NAE$\downarrow$ & 19.89 & \textbf{19.37}\\
Entity NAE$\downarrow$ & 32.42 & \textbf{26.26}\\
Formation NAE$\downarrow$ & \textbf{29.66} & 32.76\\
Correspond. NAE$\downarrow$ & 44.68 & \textbf{33.39}\\
Score$\uparrow$ & 28.84 & \textbf{40.17}\\
Group-Acc$\uparrow$ & 7.37 & \textbf{23.96}\\
Phys-Gnd$\uparrow$ & 65.96 & \textbf{78.56}\\
\bottomrule
\end{tabular}
\par}
\end{minipage}\hfill
\begin{minipage}[t]{0.45\textwidth}
\vspace{0pt}
\centering
\includegraphics[width=\linewidth]{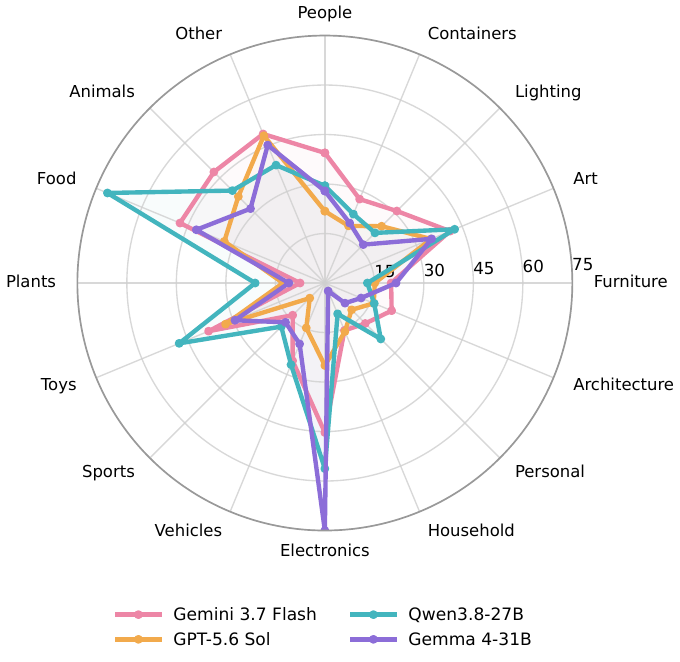}

\captionsetup{font=footnotesize,skip=3pt}
\captionof{figure}{\textbf{Target-domain performance:} group-macro Score of four representative models across 16 mutually exclusive counting-target domains.}
\label{fig:domain_analysis}
\end{minipage}
\end{minipage}

\paragraph{Impact of Question Presentation.}
Table~\ref{tab:joint_questioning} compares independent and joint presentation. Joint presentation raises Score by more than eleven points and more than triples Group-Acc, with gains concentrated in entity and correspondence questions despite slightly worse image-formation performance. Questions in the same annotation group refer to the same appearances and physical entities but request different counts. Seeing these questions together gives the model access to the full set of counting operations associated with the scene. The large reductions in entity and correspondence error show that this presentation is especially helpful for identity-dependent counts, while the rise in Group-Acc means that substantially more annotation groups are solved in full.

\paragraph{Performance across Target Domains.}
Figure~\ref{fig:domain_analysis} reveals clear differences in how stable model performance remains across counting targets. Gemini 3.7 Flash follows a relatively even profile, avoiding pronounced drops in several categories without leading every domain. Qwen3.8-27B and Gemma~4-31B reach higher peaks on selected targets but vary more across categories, and their local strengths do not transfer consistently to other targets. Several curves nevertheless rise and fall together across domains, showing that the target category systematically changes the difficulty of separating appearances and matching them to entities. The model-specific peaks further indicate that models rely on different category knowledge and identity cues.

\begin{figure*}[t]
\hyphenpenalty=10000
\exhyphenpenalty=10000
\centering

\begin{minipage}[t]{0.485\textwidth}
\vspace{0pt}
\centering
\includegraphics[
  width=\linewidth,
  height=0.4\textwidth,
  keepaspectratio
]{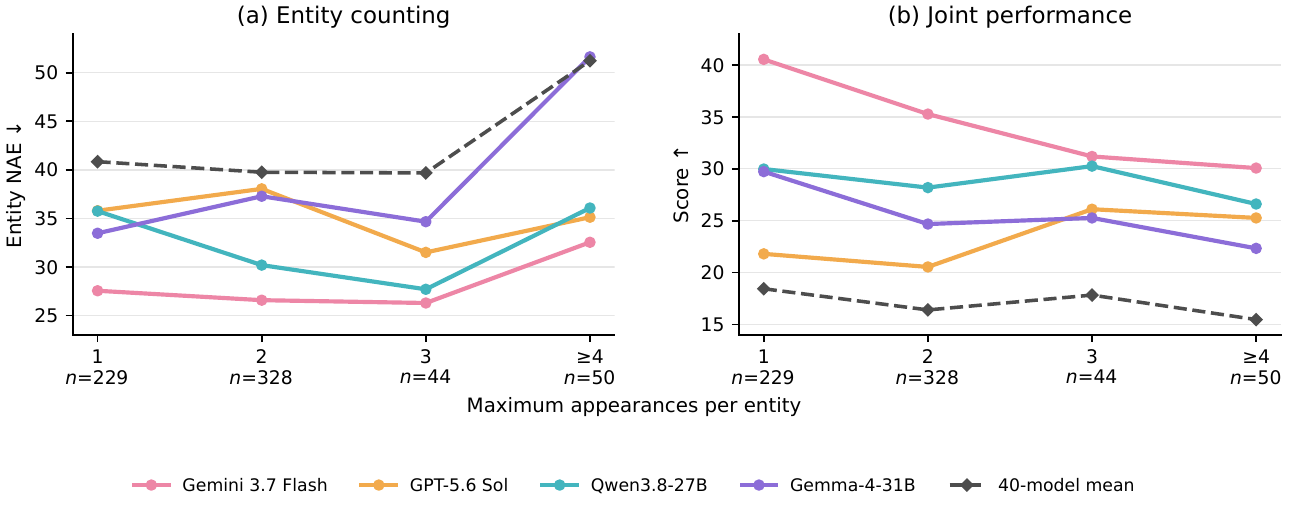}
\captionsetup{font=footnotesize,skip=3pt}
\captionof{figure}{\textbf{Performance by appearance multiplicity.} Entity NAE (a) and Score (b) grouped by the maximum number of appearances assigned to one entity; dashed lines show the equal-weight mean across all 40 models.}
\label{fig:multiplicity_analysis}
\end{minipage}
\hfill
\begin{minipage}[t]{0.485\textwidth}
\vspace{0pt}
\centering
\includegraphics[
  width=\linewidth,
  height=0.4\textwidth,
  keepaspectratio
]{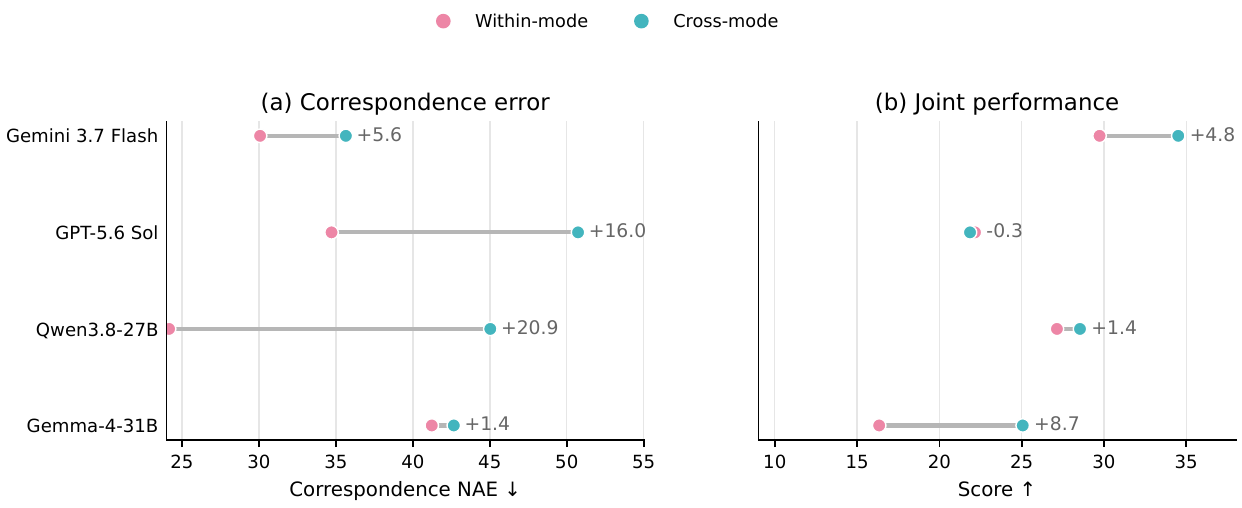}
\captionsetup{font=footnotesize,skip=3pt}
\captionof{figure}{\textbf{Within- v.s. cross-mode correspondence.} Correspondence NAE (a) and Score (b) after matching the appearance-multiplicity distributions of the two group types; annotations report the cross-mode minus within-mode difference.}
\label{fig:cross_mode_analysis}
\end{minipage}

\end{figure*}


\paragraph{Impact of Appearance Multiplicity.}
Figure~\ref{fig:multiplicity_analysis} shows that appearance multiplicity affects entity counting and joint performance at different stages. The 40-model mean entity NAE changes little from one to three appearances, then rises sharply in the four-or-more bin. Mean Score declines as multiplicity increases from one to two, partially recovers at three while remaining below the single-appearance level, and reaches its minimum at four or more. Gemini 3.7 Flash declines steadily across the four bins, whereas the other representative models fluctuate at intermediate multiplicities; nevertheless, 32 of the 40 models score lower in the highest-multiplicity bin than in the single-appearance bin. A small number of repeated appearances can therefore leave the entity count largely intact while already reducing the likelihood that appearance, entity, image-formation, and correspondence questions are all answered reliably. Each additional appearance supplies more visual evidence but also adds another correspondence that the requested entity count must resolve. These effects can offset each other at lower multiplicities; with four or more appearances, both entity counting and joint performance deteriorate.


\paragraph{Impact of Matching across Observation Modes.}
After matching the appearance-multiplicity distributions, Figure~\ref{fig:cross_mode_analysis} shows higher cross-mode correspondence error for all four representative models, with the same pattern appearing in 31 of the 40 evaluated models. Counts based on identity correspondence are thus sensitive not only to the number of observations, but also to how those observations are formed. Models can answer simpler appearance questions correctly while still missing counts that require direct, reflected, and transmitted observations to be treated as views of the same entity. Aggregate Score does not decline in parallel because direct views can simplify individual counts and offset errors in cross-mode correspondence. A competitive Score can therefore coexist with substantial errors on questions that require cross-mode identity matching.

\paragraph{Discussion.}
Across the four analyses, models improve at enumerating appearances before they reliably answer counts defined by physical identity. Higher resolution improves access to visible evidence, while repeated and cross-mode observations remain difficult; the joint-prompting experiment also improves complete-group correctness. The central challenge is therefore to organize observations around persistent physical entities rather than count them in isolation. Future evaluation can complement final counts with explicit appearance and identity predictions, separating missed evidence from incorrect grouping. For physical AI, this points toward scene understanding whose entity-level predictions remain stable across image formation, viewpoint, and time~\cite{nvidia2026cosmos3omnimodalworld}.

\section{Conclusion}
\label{sec:conclusion}
We introduced PhysCount, a benchmark for evaluating whether MLLMs distinguish visual appearances from the physical entities that produce them. By deriving 4,595 related questions from shared appearance--entity annotations, PhysCount evaluates individual counts and complete-group correctness across appearance, entity, image-formation, and correspondence questions. Experiments with 40 MLLMs show that appearance counts are more reliable than counts requiring physical-identity correspondence, particularly when entities appear repeatedly or across observation modes. Higher resolution reduces appearance error without eliminating the remaining identity-dependent errors, while joint presentation substantially increases the number of groups solved in full. These results separate two observable capabilities: enumerating visible evidence and answering counts defined by how that evidence maps to physical objects. Progress toward physical scene understanding should make entity-centered predictions stable across reflection, transmission, viewpoint, and time. Such stability would support reasoning about enduring objects and their interactions rather than treating each visual occurrence as an independent counting unit. PhysCount exposes the gap between counting what is visible and correctly counting the physical entities those observations depict.

\subsection*{AI Use Statement}
Generative AI tools assisted candidate-image screening, consistency checks of
question--answer pairs against structured annotations, language editing, and
manuscript organization. Human annotators made the final image-selection
decisions, created and reviewed the annotations and questions, and adjudicated
cases flagged during QA checking. The authors reviewed the resulting data,
text, claims, and artifacts and take full responsibility for the work.

\subsection*{Reproducibility Statement}
Section~\ref{sec:evaluation_protocol} defines the metrics and aggregation
procedure, and Section~\ref{sec:evaluation_setting} specifies the common
evaluation setting. Appendix~\ref{app:annotation} provides the scene-selection
criteria, screening and QA-audit prompts, annotation-verification procedure,
and representative examples pairing questions and answers with their structured
annotations. Appendix~\ref{app:evaluation_details} gives the
evaluation prompts, image preprocessing, inference settings, human-baseline
protocol, and answer-validation rules. Appendix~\ref{app:diagnostic_results}
reports the complete numerical results underlying the diagnostic analyses.
The complete PhysCount benchmark will be released upon acceptance.

\bibliography{iclr2027_conference}
\bibliographystyle{iclr2027_conference}

\clearpage
\appendix
\section{Scene Curation and Annotation Verification}
\label{app:annotation}
Section~\ref{sec:data_collection} describes the construction of PhysCount. This section focuses on two boundaries that require further detail: when a mirror or glass scene genuinely changes the count, and how physical identity is established when no direct appearance is visible. We then describe the QA audit and provide complete annotation examples.

\subsection{Candidate Scene Boundaries}

Candidate images are drawn from Common Canvas~\cite{Gokaslan_2024_CVPR}, GPIC~\cite{chandrasegaran2026gpicgiantpermissiveimage}, and other public collections. We retain two scene types: reflection produces a clear duplicate of a countable target, or targets behind glass overlap with salient surface reflections in a way that changes target boundaries or apparent multiplicity. A reflective surface alone is insufficient, and the target must remain reliably countable after careful inspection.

The curation criteria separate physical existence from the path of observation. Objects seen directly or through a transparent medium are treated as real objects; only appearances formed on a reflective surface are reflections. Content displayed on televisions, posters, photographs, or paintings is excluded, as are shadows, highlights, and weak reflections on wet surfaces. One entity together with a complete mirror appearance is sufficient to create meaningful appearance duplication. Ordinary transmission is not: an object merely seen through glass does not enter the benchmark unless a surface reflection also introduces duplication, overlap, or an identity relation.

Qwen3.6-35B-A3B~\cite{qwen36_35b_a3b} evaluates each candidate twice. The first pass identifies the counting target and checks countability, reflection salience, and counting ambiguity. Only candidates satisfying every condition proceed to a second judgment. The second pass receives the image and target category without the first-pass rationale, and independently verifies the duplicated or mixed reflection--transmission structure. Both records are validated as structured JSON, and a fixed conjunction of their fields determines retention.

\subsubsection{First-Pass Screening}

The first pass rejects weak or incomplete reflections, reflective content unrelated to the target, and scenes that do not support reliable human counting. Source captions serve only as retrieval cues; visible evidence determines the decision.

\begin{PromptBlock}{Stage-one screening}
You are a strict data screener for a visual counting benchmark. Make judgments
only from evidence visible in the current image. Do not rely on information in
the caption that is not visible in the image.

Your task is to find real images in which reflection or transmission creates a
clear counting structure. Prefer rejecting many candidates over retaining a
weakly relevant image.

Here, "counting ambiguity" does not require a reflection to completely deceive
the viewer. It means that reflection/transmission creates salient repetition,
superposition, partitioning, or occlusion, such that REAL and
REFLECTION/TRANSMISSION must be distinguished to count correctly.

- REAL: a real object seen directly or through a transparent medium.
- REFLECTION: an appearance reflected by a mirror, glass, water, a glossy floor,
  or a metallic surface.
- An object seen through glass remains REAL; only the reflected appearance
  superimposed on the glass surface is REFLECTION.
- Images shown on televisions, posters, photographs, paintings, or screens are
  not reflections. Shadows are not reflections either.

Set KEEP=YES only if at least one of the following paths is satisfied:
A. Duplicated appearance: a reflection contains a clearly delineated repeated
   appearance of the same category that an ordinary viewer might count twice.
B. Reflection-transmission mixture: real countable targets behind a shop
   window, glass door, display case, or similar surface significantly overlap,
   join, shift relative to, or occlude reflections on the glass surface, truly
   creating missed counts, double counts, or ambiguous target boundaries.

Both paths must satisfy all of the following:
1. The reflection is sufficiently salient, rather than a weak highlight, a
   locally blurred reflection, or an incidental background detail.
2. A genuine counting ambiguity exists.
3. After careful inspection, a human can still reliably identify the primary
   counting target.

The following structures are positive examples calibrated from human-selected
cases:
- Mirror walls, mirrored columns, or multiple mirrors copy the same person or
  object into multiple regions.
- Real people, products, mannequins, or other objects inside a shop window are
  clearly superimposed with reflected street scenes or people on the glass.
- Glass in a shop, cafe, or display case simultaneously presents countable
  interior targets and strong exterior reflections.
- A person stands before a mirror, and both the real person and the reflected
  person are complete, salient, and separately countable.
- Even when there is only one real person, retain the image if the person and a
  complete, clear mirror appearance are visible simultaneously.
- Multi-faceted mirrors, mirror installations, or highly reflective objects
  produce multiple clear copies.
- An object placed on a mirrored surface and its complete reflection form two
  unambiguous appearances.

High-value cases include:
1. Targets of the same category occur as both REAL and REFLECTION.
2. One real target produces multiple reflected appearances.
3. Real targets behind glass and reflections on the glass surface coexist.
4. Reflection/transmission causes overlap, displacement, occlusion, or joining
   that can induce double counting or missed counting.
5. The image supports separate questions about the number of real objects,
   reflected appearances, real objects seen through a transparent medium, and
   total visual appearances.
6. There are many targets, but humans can still count them reliably.

Always set KEEP=NO for the following:
- Targets are merely seen through water or glass, without a clear reflected
  duplicate of the same category.
- Reflections on wetlands, stage floors, or water are blurred, incomplete, or
  very faint.
- A reflection only indicates that a surface is reflective and would not be
  mistaken for an additional target.
- The real and reflected targets belong to different categories and there is no
  salient reflection-transmission mixture.
- The reflection is distant, tiny, or inconspicuous and does not affect an
  initial count.
- There is only a faint reflection on water, wet ground, or an ordinary floor,
  without a clear duplicated structure.
- Fish are merely seen through water or people are merely seen through glass,
  with no strong reflection superimposed on the surface.

The following must also be rejected:
- Only lights, sky, architectural textures, or abstract colors are reflected,
  with no clear countable target.
- A mirror or glass surface is present but irrelevant to counting the target.
- Targets are too small, blurred, or heavily occluded for reliable human
  counting.
- Product photographs consisting only of a mirror, glass, water, or metal.
- Illustrations, diagrams, logos, text pages, purely abstract images, collages,
  or multi-panel images.
- No single clear primary counting category can be identified.

Judgment requirements:
- Select only one most suitable primary counting category; do not mix multiple
  categories.
- countability measures whether a human can reliably identify targets one by
  one. Do not assign a high rating merely because the image appears crowded.
- reason must be concise and no longer than one sentence.

Output only the complete JSON required by the schema. Do not provide an
explanation or chain of thought.
\end{PromptBlock}

The response records the proposed target and scene type together with judgments of countability, reflection salience, and counting ambiguity. Every condition must be satisfied before verification.

\subsubsection{Independent Verification}

The verification pass reassesses whether the scene is photographic, the target is countable, and reflection creates either genuine duplication or a consequential interaction with transmission. Rejected candidates are assigned to fixed categories covering non-photographic content, mediated images, abstract reflections, irrelevant glass, unclear or uncountable targets, and structurally simple scenes.

\begin{PromptBlock}{Independent verification}
You are a high-precision reviewer for reflection/transmission visual counting
images. Inspect the image independently and do not blindly follow the first
stage.

Only decide whether the image should enter the high-quality retrieval set. Do
not generate questions or count answers.

All of the following requirements must be satisfied:
1. The image is a real photograph or a credible real-world scene, rather than
   an illustration, diagram, logo, text page, screen content, or collage.
2. There is one clear primary counting category whose targets are sufficiently
   large and visually identifiable by humans.
3. Same-category targets form clearly duplicated appearances, or real targets
   behind glass significantly overlap, join, or occlude reflections on the
   surface.
4. At first glance, the structure can genuinely cause double counting, missed
   counting, or target-boundary confusion; merely detecting glass or glare is
   insufficient.
5. The reflection does not consist only of lights, sky, architectural texture,
   color, or highlights.
6. Targets are not so small, blurred, or occluded that reliable counting is
   impossible.

Decision standard:
- verified_keep=YES requires either clear same-category duplicated appearances,
  or a salient mixture between transmitted targets in a shop window/glass case
  and surface reflections. The reflection must be salient, a real counting risk
  must exist, and the targets must remain identifiable after careful inspection.
- Reject blurred, faint, incomplete, tiny, or background reflections, as well
  as reflections that an ordinary viewer would not mistake for an extra target.
- Reject ordinary underwater fish, objects merely viewed through glass, weak
  wet-ground reflections, and weak stage-floor reflections, unless a salient
  reflected duplicate of the same category is also present.

A single real target together with its complete and clear mirror appearance may
enter the high-quality retrieval set. Ordinary transmission without
superimposed reflection must still be rejected. Write only one sentence in
reason.

Output only complete and valid JSON.
\end{PromptBlock}

\subsubsection{Final Selection}

The final decision uses no additional model score. A retained image must satisfy every first-pass condition and receive an affirmative verification decision with high countability, counting value, and reflection salience, confirmed counting ambiguity, and no false-positive type.

Ordinary \texttt{transmission} does not pass this rule even when the transmitted targets are clear. Through-glass appearances remain eligible inside a mixed reflection--transmission scene. This distinction ensures that every retained image contains an image-formation effect that changes the correspondence between visible appearances and physical entities.

\subsection{Annotation Verification}

A physical entity need not have a direct appearance in the image. Several mirror-only appearances may belong to one entity, while direct, reflected, and through-glass appearances may all refer to the same object. Multi-mirror identities are recorded as complete same-instance groups rather than disconnected pairwise links. When an image contains several suitable target categories, each category forms a separate annotation group.

The first annotator marks appearance locations and modes, groups appearances by physical identity, and writes the questions and answers from the image. A second annotator checks the appearance locations, image-formation labels, and identity groups against the source image. GPT-5.6 Sol then recomputes each answer from the verified annotation and checks whether the question meaning agrees with that calculation. A third annotator resolves every flagged case.

Appearance questions count annotated observations; entity questions count physical objects or identity groups; image-formation questions select observations by mode; and correspondence questions compute intersections, exclusions, and multiplicities over identity groups. This shared annotation makes every answer auditable from the same scene record.

\subsubsection{Question--Answer Consistency Audit}

For each question, the auditor identifies the requested counting unit, selects the relevant appearances or identity groups, and derives the expected answer through an explicit set operation. The audit covers total and mode-specific appearances, distinct entities, entities defined by their available appearance types, reflection multiplicity, and counterfactual changes to an image-formation path.

The derived answer is compared directly with the recorded QA. Agreement is marked \texttt{correct}; a supported question with a conflicting answer is marked \texttt{contradicted}. Ambiguous questions, image--annotation conflicts, and insufficient evidence are sent to manual review, while questions requiring an unannotated attribute are marked \texttt{unsupported}. A semantically correct but unnatural question may be rewritten without changing its answer and is marked \texttt{revised}.

\begin{PromptBlock}{Question--answer consistency audit}
You are auditing one visual counting annotation group for a benchmark about
physical objects, mirror reflections, and transparent-media views.

You receive:
1. The source image.
2. An optional overlay image showing annotated points, labels, identity groups,
   and relations.
3. The structured annotation for one group.
4. Existing question-answer pairs written for this unit.

Your task is to independently verify every QA using the image and annotation.
Do not invent objects, relations, image-formation paths, or quantities that
are not supported by the supplied evidence.

IMPORTANT CONCEPTS

- An appearance is one annotated visual occurrence in the image.
  A physical-space appearance, a mirror reflection appearance, and a
  transmission appearance are all appearances.

- A physical entity is a distinct real-world object after merging all
  appearances joined by the recorded identity links.

- A physical-space appearance is NOT automatically the same as a distinct
  physical entity. A real object may be visible only in a mirror or through
  glass; conversely, one entity may create multiple reflection appearances.

- The recorded identity grouping has priority when deciding whether
  appearances belong to the same object.

- Labels indicate image-formation mode:
  - direct: an appearance occupying physical space
  - reflection: an appearance formed by a mirror or reflective surface
  - through-glass: an appearance seen through a transparent medium

- For questions about "different", "distinct", "actual objects in the scene",
  or wording that explicitly merges an object and its reflections, count
  physical entities, not appearances.

- For questions about what "occupies physical space", count only
  physical-space appearances. Do not silently deduplicate them into entities.

- For questions about objects visible only in a mirror, count identity groups
  that have reflection appearance(s) but no physical-space appearance.
  For objects visible only physically, count identity groups that have
  physical-space appearance(s) but no reflection appearance.
  For objects visible both physically and in a mirror, count identity groups
  that have both.

- For questions about the maximum number of mirror images of one object,
  compute the maximum number of reflection appearances linked to any one
  identity group.

AUDIT PROCEDURE FOR EACH QA

1. Read the question literally and classify its count target.
2. Identify the exact relevant appearances or entity groups in the annotation.
3. Compute the expected numerical answer from the annotation.
4. Check the source image and overlay for obvious contradictions, ambiguity,
   annotation mistakes, or insufficient evidence.
5. Compare the expected answer with the existing answer.
6. Improve the question wording only if needed. Keep it natural, precise, and
   a counting question. Do not reveal the solution method or leak hidden
   annotation facts into the wording.
7. Return a verdict and an explicit derivation.

VERDICT RULES

- correct:
  The question is natural enough, the annotation supports the required count, and
  the existing answer matches the derived answer.

- revised:
  The question wording requires revision, but its meaning is supported and the
  existing numerical answer matches the derived answer.

- contradicted:
  The question is answerable from the annotation, but its existing answer conflicts
  with the derived answer.

- needs_manual_review:
  The count cannot be confirmed safely because the image, overlay, question
  semantics, or annotation creates genuine ambiguity or conflict.

- unsupported:
  The question requires information not represented by the annotation, so the
  answer cannot be calculated reliably.

OUTPUT REQUIREMENTS

For each question, return the verdict, the verified answer, any revised
wording, the counting target, and an explicit derivation. For example:

"Entities with direct appearances: {E1, E2, E3}.
Entities with both direct and reflected appearances: {E1, E3}.
Therefore, only E2 is visible directly but not by reflection, so the answer is 1."

Name the relevant appearances, entity groups, relations, or set operation;
do not use vague claims such as "the annotation indicates this".
\end{PromptBlock}

\subsection{Worked Annotation and QA Examples}

\begin{figure}[!htbp]
\centering
\includegraphics[width=0.88\linewidth,trim=0 94 0 0,clip]{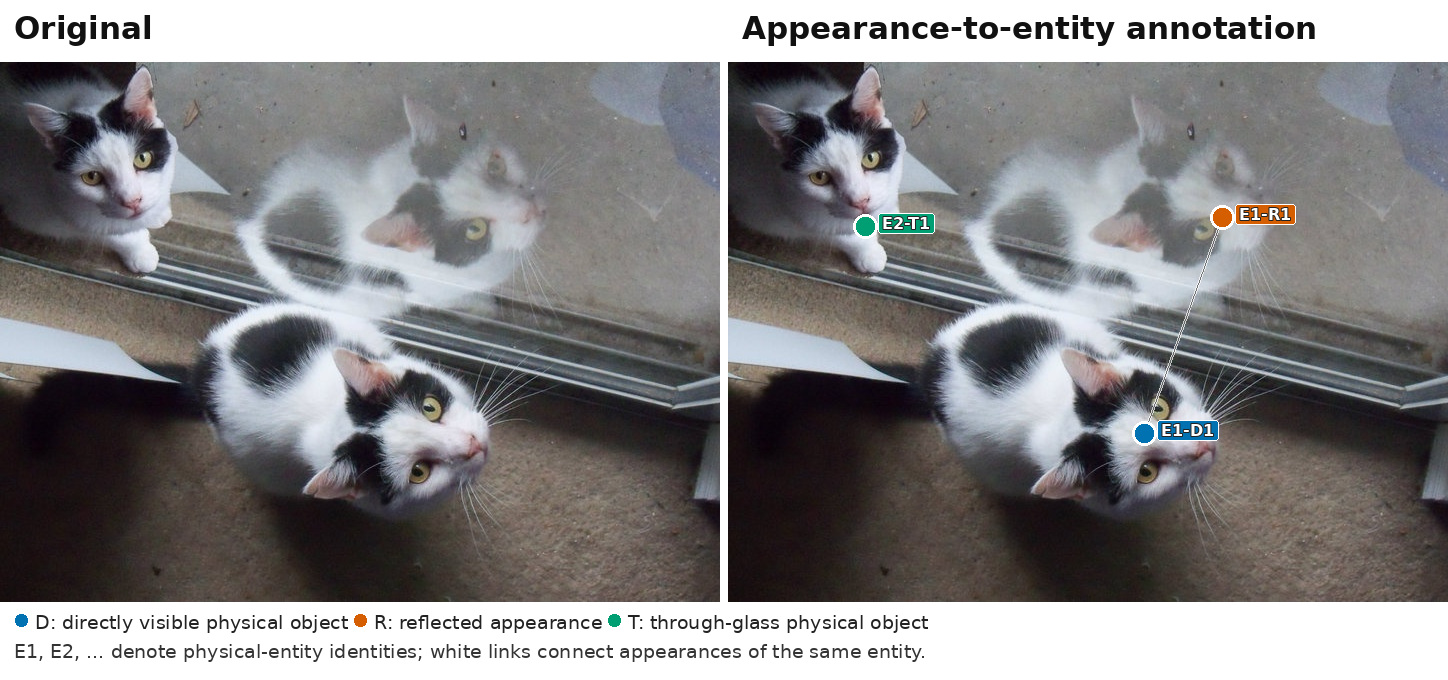}
\vspace{0.35em}

{\small
\modechip{physdirect}{D$\times$1}\quad
\modechip{physreflect}{R$\times$1}\quad
\modechip{phystransmit}{T$\times$1}
$\quad\Longrightarrow\quad E_1=\{D_1,R_1\},\;E_2=\{T_1\}$\par}
\vspace{0.35em}
{\scriptsize\setlength{\tabcolsep}{4pt}
\begin{tabular}{@{}c p{0.82\linewidth} c@{}}
\toprule
\textbf{Category} & \textbf{Question} & \textbf{Answer} \\
\midrule
A & Counting both the cats physically present in the photo, including any seen through the glass door, and the cat reflections on the glass door, how many times do cats appear in total? & 3 \\
E & How many different cats are actually in the scene photographed? & 2 \\
I & How many cat reflections are visible on the glass door? & 1 \\
I & How many cats can be seen through the glass door that extends horizontally across the image? & 1 \\
C & Among the cats on the same side of the glass door as the photographer, how many do not form a reflection on the glass door? & 0 \\
C & What is the maximum number of reflections of a single cat on the glass door? & 1 \\
\bottomrule
\end{tabular}}
\caption{\textbf{Example 1: reflection and transmission of the same target category.} Three cat appearances form two physical identities: one identity is observed directly and by reflection, while the other is seen through glass. The table reports every question derived from these annotations.}
\label{fig:qa_example_1}
\end{figure}

\begin{figure}[!htbp]
\centering
\includegraphics[width=0.82\linewidth,trim=0 94 0 0,clip]{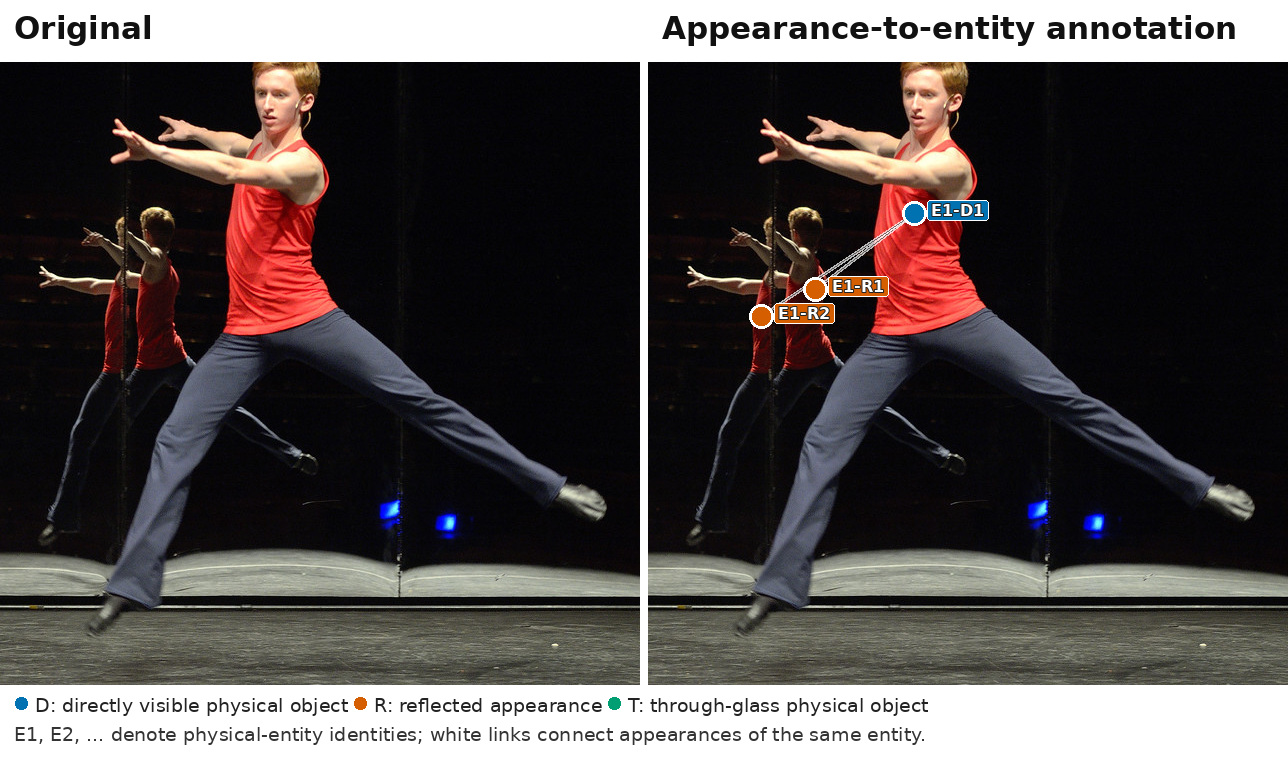}
\vspace{0.35em}

{\small
\modechip{physdirect}{D$\times$1}\quad
\modechip{physreflect}{R$\times$2}
$\quad\Longrightarrow\quad E_1=\{D_1,R_1,R_2\}$\par}
\vspace{0.35em}
{\scriptsize\setlength{\tabcolsep}{4pt}
\begin{tabular}{@{}c p{0.82\linewidth} c@{}}
\toprule
\textbf{Category} & \textbf{Question} & \textbf{Answer} \\
\midrule
A & How many visible appearances of people are there in total in the photo, including the people on the stage and their reflections in the mirror? & 3 \\
E & How many people physically occupy space on the stage, excluding their reflections in the mirror? & 1 \\
E & How many different people are actually in the scene captured in this photo? & 1 \\
I & How many people's reflections are shown in the mirror in total? & 2 \\
C & How many people appear only in the physical space of the stage, with no reflection in the mirror? & 0 \\
C & How many people are visible only as reflections in the mirror, with no corresponding person visible in the physical space of the stage? & 0 \\
C & How many people appear both in the physical space of the stage and in the mirror? & 1 \\
C & What is the maximum number of reflections of the same person shown in the mirror? & 2 \\
\bottomrule
\end{tabular}}
\caption{\textbf{Example 2: one person with two reflected appearances.} The direct appearance and both mirror appearances share one identity. The questions contrast appearance enumeration with physical occupancy, identity, and reflection multiplicity.}
\label{fig:qa_example_2}
\end{figure}

\begin{figure}[!htbp]
\centering
\includegraphics[width=0.88\linewidth,trim=0 94 0 0,clip]{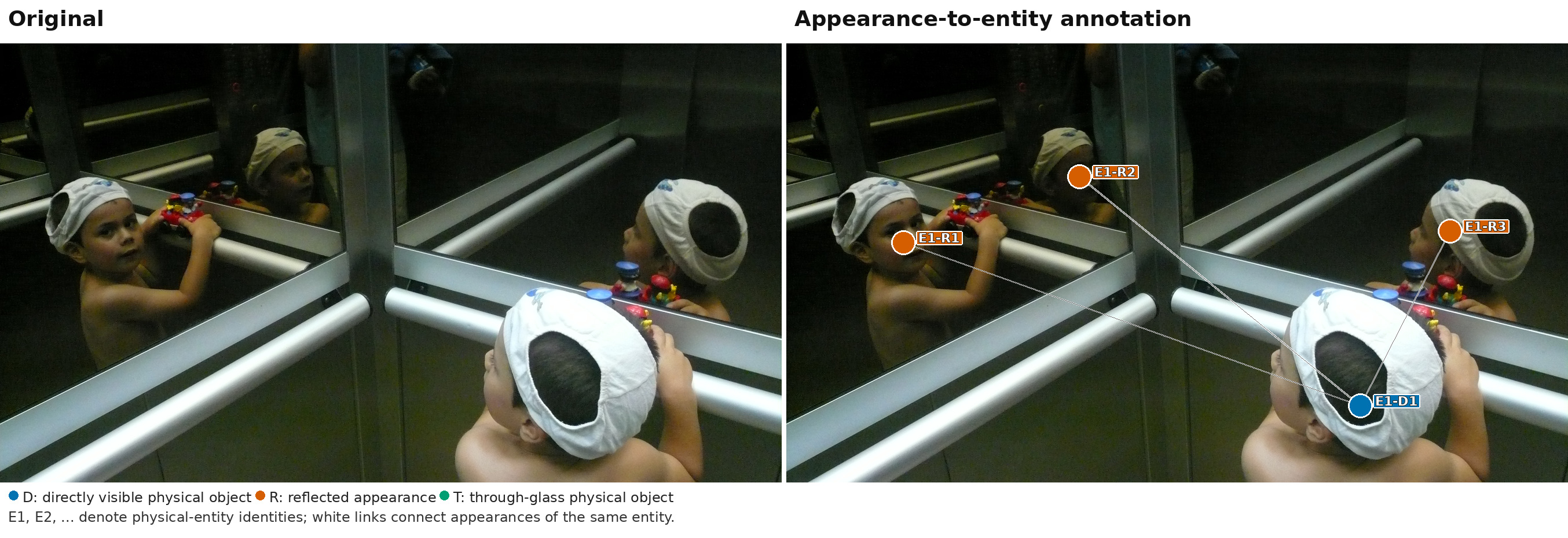}
\vspace{0.35em}

{\small
\modechip{physdirect}{D$\times$1}\quad
\modechip{physreflect}{R$\times$3}
$\quad\Longrightarrow\quad E_1=\{D_1,R_1,R_2,R_3\}$\par}
\vspace{0.35em}
{\scriptsize\setlength{\tabcolsep}{4pt}
\begin{tabular}{@{}c p{0.82\linewidth} c@{}}
\toprule
\textbf{Category} & \textbf{Question} & \textbf{Answer} \\
\midrule
A & How many visible person appearances are there in the photo in total? & 4 \\
E & How many people are physically present in the photo? & 1 \\
E & Not counting multiple images of the same person as different people, how many distinct people are there in the photo? & 1 \\
I & How many mirror images of people are visible in the photo? & 3 \\
C & How many people are both physically present and visible in the mirror? & 1 \\
C & How many people are visible only as reflections in the mirror, with no corresponding person physically present in the photo? & 0 \\
C & How many people have multiple mirror images? & 1 \\
C & What is the maximum number of mirror images of the same person that appear in the photo at the same time? & 3 \\
\bottomrule
\end{tabular}}
\caption{\textbf{Example 3: one person distributed across three mirror regions.} Four appearances map to one physical identity, enabling questions about total appearances, reflected subsets, physical--reflection overlap, and maximum multiplicity.}
\label{fig:qa_example_3}
\end{figure}

\begin{figure}[!htbp]
\centering
\includegraphics[width=0.88\linewidth,trim=0 94 0 0,clip]{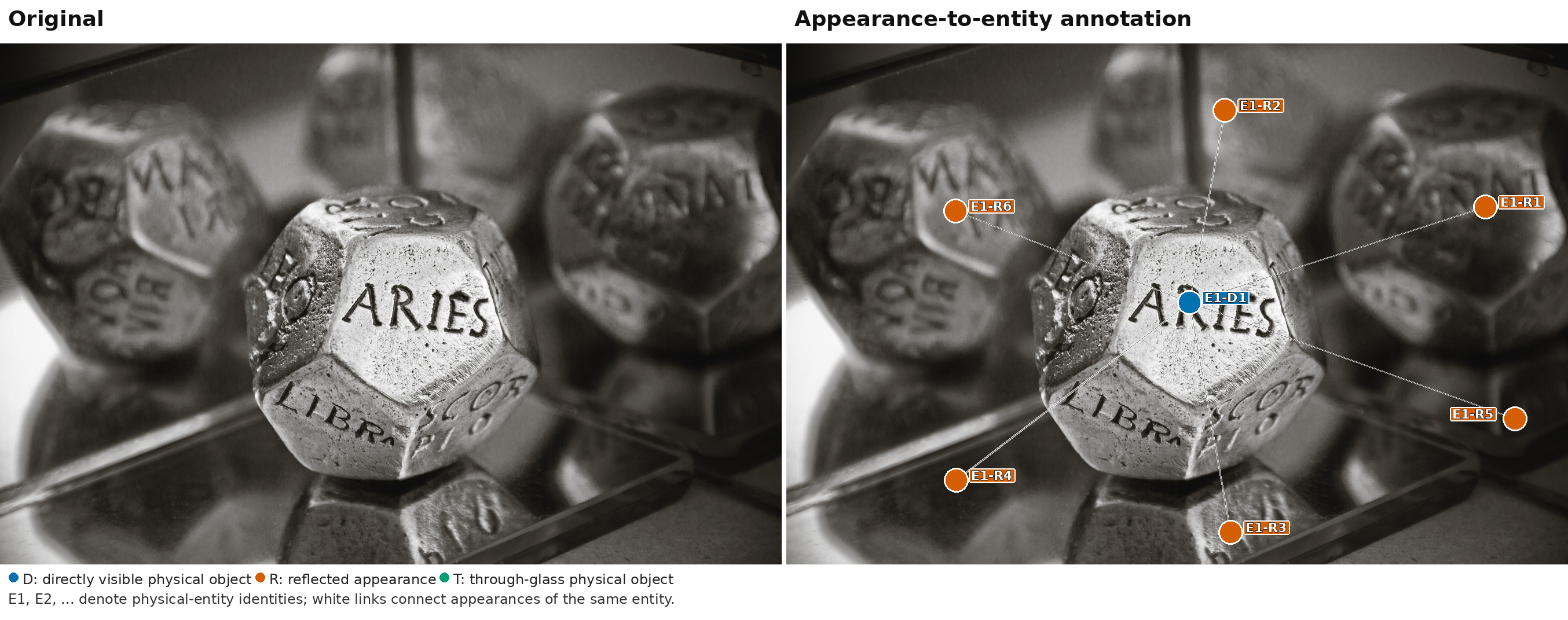}
\vspace{0.35em}

{\small
\modechip{physdirect}{D$\times$1}\quad
\modechip{physreflect}{R$\times$6}
$\quad\Longrightarrow\quad E_1=\{D_1,R_1,\ldots,R_6\}$\par}
\vspace{0.35em}
{\scriptsize\setlength{\tabcolsep}{4pt}
\begin{tabular}{@{}c p{0.82\linewidth} c@{}}
\toprule
\textbf{Category} & \textbf{Question} & \textbf{Answer} \\
\midrule
A & Counting both the dice physically present in the photo and the mirror reflections, how many dice appear in total? & 7 \\
E & How many dice are physically present in the photo? & 1 \\
E & How many distinct dice are actually in the scene captured by this photo? & 1 \\
I & How many mirror reflections of dice are there in the photo? & 6 \\
C & How many dice have more than one reflection in the mirror? & 1 \\
C & How many dice are physically present in the photo but do not appear in the mirror? & 0 \\
C & What is the maximum number of reflections formed in the mirror by the same die? & 6 \\
\bottomrule
\end{tabular}}
\caption{\textbf{Example 4: high reflection multiplicity from a single object.} One physical die produces six reflected appearances. The resulting questions separate the seven visible appearances from one physical identity and its reflection multiplicity.}
\label{fig:qa_example_4}
\end{figure}

\begin{figure}[!htbp]
\centering
\includegraphics[width=0.88\linewidth,trim=0 94 0 0,clip]{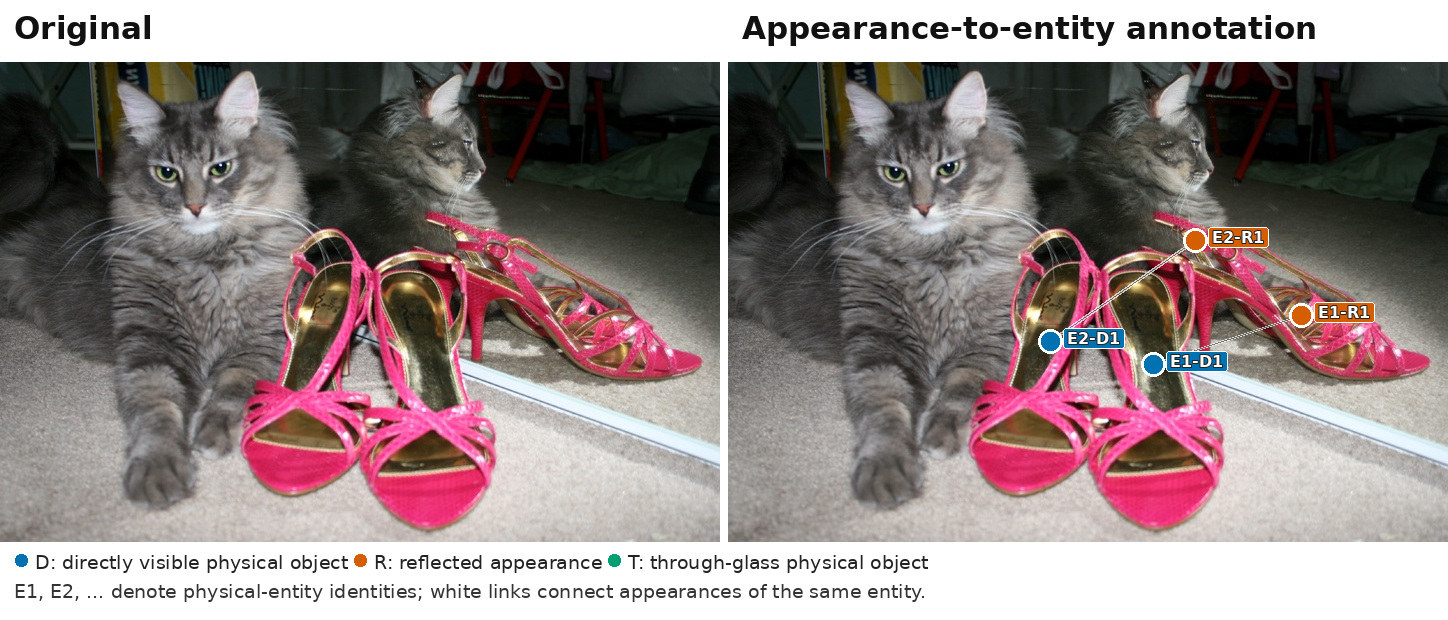}
\vspace{0.35em}

{\small
\modechip{physdirect}{D$\times$2}\quad
\modechip{physreflect}{R$\times$2}
$\quad\Longrightarrow\quad E_1=\{D_1,R_1\},\;E_2=\{D_2,R_2\}$\par}
\vspace{0.35em}
{\scriptsize\setlength{\tabcolsep}{4pt}
\begin{tabular}{@{}c p{0.82\linewidth} c@{}}
\toprule
\textbf{Category} & \textbf{Question} & \textbf{Answer} \\
\midrule
A & How many shoes appear in total, counting both the shoes occupying physical space in the photo and their reflections in the mirror on the right? & 4 \\
E & How many shoes in the photo occupy physical space? & 2 \\
E & How many different shoes are actually in the scene captured in this photo? & 2 \\
I & How many shoes are reflected in the mirror on the right? & 2 \\
C & How many shoes occupy physical space but have no corresponding reflection in the mirror on the right? & 0 \\
C & How many shoes appear only in the mirror on the right, with no corresponding shoes occupying physical space in the photo? & 0 \\
C & How many shoes both occupy physical space in the photo and appear as reflections in the mirror on the right? & 2 \\
C & What is the maximum number of times the same shoe appears as a reflection in the mirror on the right? & 1 \\
\bottomrule
\end{tabular}}
\caption{\textbf{Example 5: two physical objects paired with their mirror appearances.} Four shoe appearances form two identity groups. The questions require two independent direct--reflection correspondences rather than a single global count.}
\label{fig:qa_example_5}
\end{figure}

\begin{figure}[!htbp]
\centering
\includegraphics[width=0.88\linewidth,trim=0 94 0 0,clip]{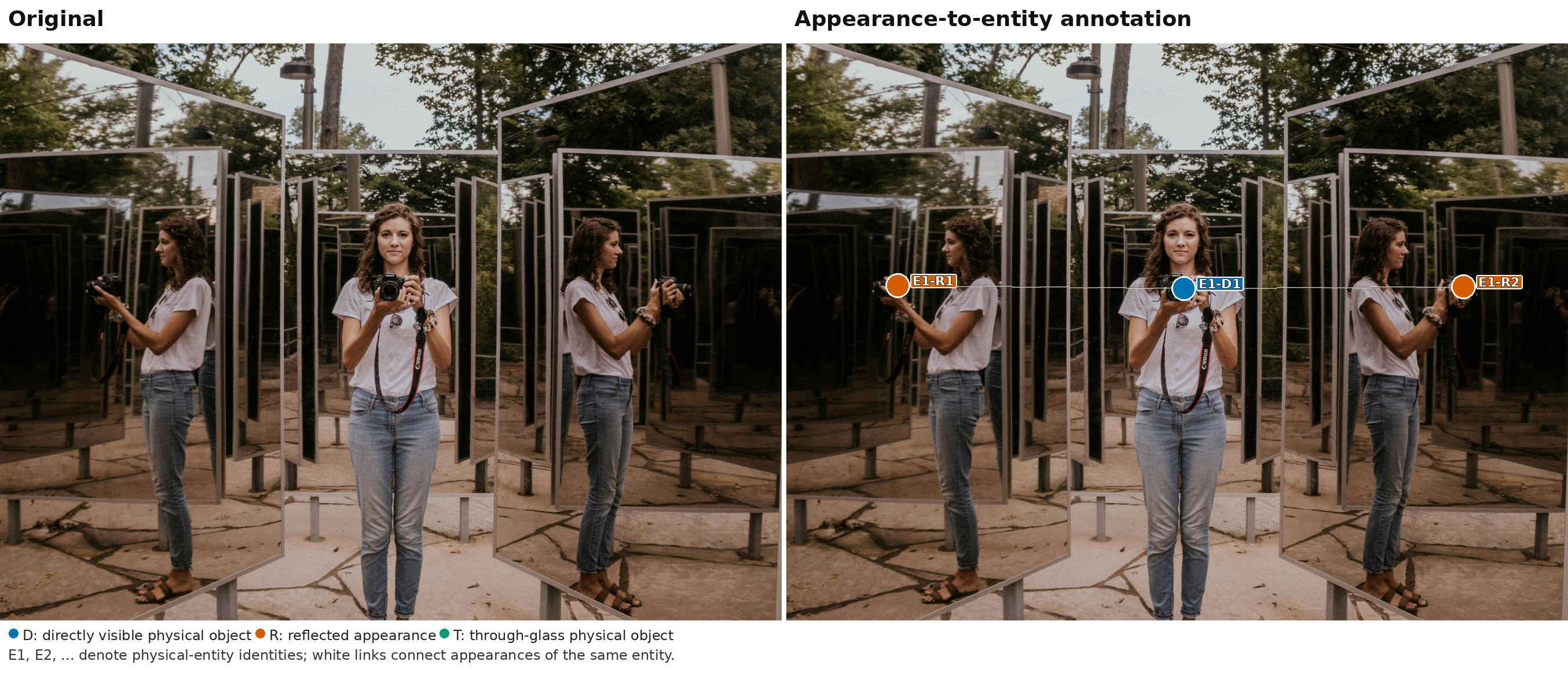}
\vspace{0.35em}

{\small
\modechip{physdirect}{D$\times$1}\quad
\modechip{physreflect}{R$\times$2}
$\quad\Longrightarrow\quad E_1=\{D_1,R_1,R_2\}$\par}
\vspace{0.35em}
{\scriptsize\setlength{\tabcolsep}{4pt}
\begin{tabular}{@{}c p{0.82\linewidth} c@{}}
\toprule
\textbf{Category} & \textbf{Question} & \textbf{Answer} \\
\midrule
A & How many cameras appear in total in the photo? & 3 \\
E & How many cameras occupy physical positions in the photo? & 1 \\
E & How many different cameras are in the photo? & 1 \\
I & How many cameras appear in the mirror in the photo? & 2 \\
C & How many cameras appear both in a physical position and in the mirror? & 1 \\
C & What is the maximum number of mirror images that any single camera has in the photo at the same time? & 2 \\
\bottomrule
\end{tabular}}
\caption{\textbf{Example 6: a direct camera repeated across two mirrors.} Three camera appearances belong to one physical identity. The questions distinguish the direct camera, the reflected subset, their shared identity, and reflection multiplicity.}
\label{fig:qa_example_6}
\end{figure}

Figures~\ref{fig:qa_example_1}--\ref{fig:qa_example_6} show all questions from six reviewed annotation groups. Each figure pairs the source image with its appearance--entity overlay. Blue \modechip{physdirect}{D} nodes mark direct physical-space appearances, orange \modechip{physreflect}{R} nodes mark reflected appearances, and green \modechip{phystransmit}{T} nodes mark appearances seen through transparent media. Shared entity identifiers and connecting lines show which appearances depict the same object. The compact diagram below each image summarizes these identities, and the table gives every QA and its verified answer. Category A counts all visual appearances, E counts objects occupying physical space or distinct physical entities, I counts appearances by image-formation mode, and C queries relations between appearances and identity groups.

\clearpage
\section{Evaluation Details and Reproducibility}
\label{app:evaluation_details}

\subsection{Prompting Protocols}

\paragraph{Independent QA.} Each image--question pair is evaluated in a separate model call. The image and the following prompt are supplied together in one user message:

\begin{PromptBlock}{Independent question answering}
Answer the following counting question about this image. Carefully distinguish
visual appearances from physical entities, including appearances formed by
reflections or seen through glass, and use appearance-entity correspondences
when required by the question. Follow the counting target specified in the
question. Return exactly one non-negative integer, with no labels or prose.

Question: {question}
\end{PromptBlock}

Questions from the same annotation group are answered without access to one another. No additional system instruction is used.

\paragraph{Group-level joint prompting.} All factual questions from one annotation group are placed in a single prompt with one shared image, and the answers follow the question order.

\begin{PromptBlock}{Group-level joint questioning}
Answer the following counting questions about this image. Carefully distinguish
visual appearances from physical entities, including appearances formed by
reflections or seen through glass, and use appearance-entity correspondences
when required by each question. Follow the counting target specified in each
question. Return exactly one JSON array of non-negative integers in the same
order as the {N} questions, with no labels or prose.

1. {question_1}
2. {question_2}
...
N. {question_N}
\end{PromptBlock}

The two conditions use the same task instructions, model configuration, and image preprocessing. They differ in whether questions from an annotation group are supplied separately or together and, correspondingly, whether the response is one integer or an ordered integer array.

\subsection{Image and Inference Settings}

Images retain their native resolution up to the specified long-edge limit $L$; larger images are resized isotropically without cropping or upsampling. The resolution study uses $L\in\{512,1024,2048,4096\}$, and all other experiments use $L=4096$. For each model, we select the strongest reasoning or thinking option exposed by the evaluated backend: this is \texttt{high} when that is the highest available level, \texttt{xhigh} when supported, and the model's native thinking mode when no graded effort control is provided. These backend-specific labels describe the selected operating mode and are not assumed to be numerically equivalent. We cap generation at 32,768 tokens where supported. Temperature 1.0, \texttt{top\_p=0.95}, and \texttt{top\_k=20} are applied only when the backend exposes the corresponding controls; unsupported fields are omitted.

The human baseline follows the independent-QA setting. Five participants each answer the complete set of 4,595 questions, receiving the same image and question shown to a model on every trial. We compute all metrics separately for each participant and report their arithmetic mean.

\subsection{Answer Validation}

Independent evaluation accepts one non-negative integer; joint evaluation accepts an integer array aligned with the question order. Missing, malformed, negative, or extreme predictions are treated as invalid: they receive maximal NAE and fail exact-match metrics, but are excluded from raw-count MAE.

\section{Expanded Diagnostic Results}
\label{app:diagnostic_results}

This section provides the numerical results behind the diagnostic analyses in
Section~\ref{sec:detailed_analysis}. All values follow the same group-macro evaluation protocol as
the main results.

\subsection{Performance across Counting Targets}

The target-domain analysis in Figure~\ref{fig:domain_analysis} shows that object recognition and
physical-entity reasoning remain coupled. A model that handles people reliably
may still fail on compact manufactured objects, furniture, or irregular natural
targets, and these shifts are not simple rescalings of overall performance.
Qwen3.8-27B is particularly strong on several compact categories, Gemini 3.7
Flash is more stable across human-like and animal targets, and Gemma~4-31B
peaks on electronics. Table~\ref{tab:domain_scores_full} reports the
corresponding Scores for all 40 models.

\subsection{Appearance Multiplicity}

Appearance multiplicity is the largest number of annotated appearances assigned to one physical entity in a group. Performance changes little across the first three bins, then breaks when an entity generates at least four appearances: mean Entity NAE rises, mean Score falls to its minimum, and 32 of 40 models score below their single-appearance result. The full trajectories in Table~\ref{tab:multiplicity_scores_full} show that the difficulty comes from maintaining a compatible identity assignment across many observations, rather than from a uniform increase in raw count.

\subsection{Within- and Cross-Mode Correspondence}

The within-/cross-mode comparison retains groups with repeated appearances and standardizes both conditions to the same multiplicity distribution. Within-mode identities repeat inside one observation mode; cross-mode identities span at least two of direct, reflected, and through-glass observation. Table~\ref{tab:cross_mode_scores_full} shows higher cross-mode correspondence error for 31 of 40 models. Score follows a different pattern because direct views support detection and enumeration even when matching them to indirect appearances remains difficult. The comparison therefore separates access to a visual anchor from the ability to preserve identity across image-formation paths.

\begin{table}[!ht]
\centering
\caption{\textbf{Performance across counting-target domains.} Each entry is the group-macro Score (\%) for one counting-target domain.}
\label{tab:domain_scores_full}
\resizebox{\linewidth}{!}{%
\begin{tabular}{lrrrrrrrrrrrrrrrr}
\toprule
Model & People & Containers & Lighting & Art & Furniture & Architecture & Personal & Household & Electronics & Vehicles & Sports & Toys & Plants & Food & Animals & Other \\
\midrule
Gemini 3.7 Flash & 39.43 & 27.53 & 30.85 & 41.06 & 20.09 & 21.91 & 17.33 & 15.54 & 45.31 & 25.51 & 13.78 & 38.13 & 7.51 & 47.51 & 47.56 & 48.89 \\
Muse-Glimmer-30B & 30.16 & 23.31 & 26.72 & 34.90 & 27.08 & 26.96 & 17.12 & 20.59 & 56.25 & 22.04 & 23.84 & 39.23 & 20.65 & 41.39 & 49.34 & 29.19 \\
Qwen3.8-27B & 29.47 & 22.60 & 21.48 & 42.57 & 12.88 & 16.19 & 23.97 & 10.10 & 56.25 & 26.78 & 18.68 & 47.78 & 21.11 & 71.31 & 39.61 & 38.68 \\
Qwen3.5-397B-A17B & 28.57 & 20.44 & 22.68 & 40.11 & 25.75 & 29.82 & 29.77 & 10.56 & 27.08 & 19.15 & 15.68 & 34.57 & 24.26 & 48.07 & 48.26 & 25.35 \\
Gemma 4-31B & 27.88 & 19.67 & 16.46 & 35.00 & 21.57 & 11.89 & 8.63 & 2.62 & 75.00 & 19.98 & 16.85 & 29.48 & 10.90 & 42.15 & 31.87 & 45.24 \\
Qwen3.5-27B & 27.11 & 19.04 & 21.06 & 42.99 & 29.98 & 18.25 & 16.47 & 14.72 & 20.31 & 25.10 & 20.87 & 34.59 & 10.23 & 28.75 & 27.54 & 35.51 \\
Qwen3.5-122B-A10B & 27.24 & 20.90 & 21.40 & 40.81 & 22.07 & 13.06 & 7.66 & 9.85 & 25.00 & 19.90 & 11.26 & 41.45 & 11.82 & 45.11 & 29.61 & 34.26 \\
Qwen3.5-35B-A3B & 27.24 & 17.79 & 19.91 & 34.48 & 19.30 & 11.93 & 19.94 & 14.04 & 31.25 & 18.12 & 8.45 & 34.01 & 9.49 & 35.81 & 27.42 & 51.02 \\
Gemma 4-26B-A4B & 24.87 & 19.67 & 6.42 & 34.09 & 18.81 & 14.49 & 8.49 & 14.25 & 37.50 & 23.95 & 9.98 & 38.24 & 5.87 & 39.48 & 26.38 & 55.99 \\
Qwen3.5-9B & 24.45 & 15.55 & 21.56 & 34.52 & 21.41 & 13.32 & 19.61 & 9.10 & 43.75 & 18.99 & 8.17 & 30.02 & 4.36 & 43.14 & 32.64 & 25.55 \\
GPT-5.6 Sol & 21.78 & 18.73 & 24.34 & 34.84 & 15.13 & 16.04 & 11.48 & 15.90 & 25.00 & 14.72 & 6.47 & 32.57 & 12.68 & 32.98 & 37.07 & 48.33 \\
InternVL3.5-241B-A28B & 21.24 & 14.76 & 23.80 & 24.58 & 11.64 & 9.56 & 17.00 & 10.53 & 12.50 & 24.02 & 25.73 & 32.95 & 11.39 & 39.28 & 46.71 & 17.82 \\
DeepSeek-V4-Flash-Vision & 20.90 & 16.49 & 22.48 & 25.08 & 20.96 & 16.35 & 4.97 & 20.50 & 37.50 & 19.37 & 6.00 & 20.61 & 9.26 & 36.37 & 36.58 & 33.75 \\
GPT-5.6 Terra & 20.85 & 17.57 & 19.08 & 19.90 & 16.87 & 19.00 & 16.99 & 6.46 & 51.04 & 20.00 & 11.48 & 14.93 & 12.88 & 22.78 & 30.43 & 41.94 \\
Gemma 4-12B & 19.28 & 16.44 & 16.00 & 23.37 & 22.75 & 10.88 & 9.01 & 12.90 & 15.62 & 20.57 & 2.38 & 33.56 & 5.15 & 40.49 & 22.53 & 28.16 \\
Qwen3.5-4B & 19.06 & 15.27 & 17.91 & 28.31 & 18.75 & 8.90 & 16.15 & 3.09 & 4.69 & 22.17 & 7.84 & 29.65 & 3.14 & 25.83 & 23.48 & 17.21 \\
InternVL3.5-30B-A3B & 18.25 & 11.61 & 20.44 & 23.81 & 10.11 & 8.18 & 16.92 & 6.92 & 12.50 & 26.12 & 19.84 & 17.86 & 13.73 & 26.73 & 21.84 & 13.98 \\
MiniMax-M3-428B-A23B & 17.93 & 11.22 & 12.37 & 22.43 & 12.29 & 8.13 & 14.16 & 23.52 & 12.50 & 12.94 & 21.33 & 17.80 & 6.00 & 19.82 & 31.21 & 44.79 \\
GPT-5.6 Luna & 17.28 & 11.53 & 16.29 & 20.24 & 10.19 & 17.37 & 12.90 & 13.49 & 0.00 & 18.38 & 2.25 & 24.30 & 11.68 & 40.34 & 26.89 & 24.44 \\
InternVL3.5-38B & 15.96 & 14.69 & 6.78 & 22.51 & 16.88 & 8.84 & 20.69 & 4.57 & 12.50 & 12.73 & 13.73 & 19.74 & 3.06 & 24.17 & 18.57 & 22.47 \\
InternVL3.5-14B & 15.49 & 11.14 & 12.24 & 23.80 & 9.62 & 10.94 & 6.91 & 9.91 & 6.25 & 19.78 & 23.41 & 17.40 & 12.81 & 13.40 & 20.59 & 27.78 \\
Molmo2-4B & 15.35 & 11.75 & 6.49 & 18.60 & 13.85 & 12.90 & 5.11 & 3.47 & 31.25 & 20.50 & 9.25 & 19.98 & 4.74 & 11.07 & 31.83 & 24.57 \\
Ovis2.6-80B-A3B & 15.39 & 13.36 & 12.34 & 11.66 & 17.12 & 9.27 & 6.68 & 6.90 & 6.25 & 15.59 & 4.05 & 21.30 & 5.95 & 15.44 & 36.06 & 12.07 \\
MiniCPM-V 4.5 & 14.98 & 10.70 & 8.21 & 22.25 & 11.03 & 10.56 & 16.31 & 14.61 & 43.75 & 11.31 & 20.43 & 13.33 & 10.56 & 18.62 & 27.27 & 22.63 \\
Cosmos3-Super & 15.01 & 9.19 & 16.08 & 30.66 & 9.87 & 8.14 & 13.25 & 10.84 & 18.75 & 16.79 & 5.48 & 12.27 & 7.40 & 19.89 & 15.79 & 19.45 \\
InternVL3.5-8B & 15.07 & 6.59 & 10.37 & 16.07 & 19.48 & 10.67 & 16.81 & 1.35 & 6.25 & 15.41 & 6.38 & 13.51 & 22.37 & 10.42 & 11.28 & 28.05 \\
Molmo2-8B & 13.47 & 11.39 & 7.13 & 11.29 & 13.59 & 12.89 & 3.92 & 6.12 & 3.12 & 19.89 & 4.47 & 20.36 & 1.45 & 16.13 & 27.56 & 19.82 \\
ERNIE 4.5 VL-28B-A3B & 13.92 & 8.38 & 5.13 & 19.83 & 14.35 & 7.42 & 8.51 & 6.45 & 0.00 & 4.48 & 14.14 & 21.70 & 3.08 & 29.16 & 19.73 & 11.15 \\
MiniCPM-V 4.6 & 12.93 & 7.55 & 11.36 & 15.13 & 7.85 & 8.24 & 21.31 & 10.22 & 4.69 & 11.79 & 7.50 & 10.06 & 4.82 & 29.54 & 11.50 & 15.65 \\
Cosmos3-Nano & 12.36 & 8.38 & 7.16 & 22.81 & 7.53 & 6.50 & 19.38 & 4.11 & 9.38 & 16.74 & 5.37 & 11.18 & 1.15 & 19.02 & 17.52 & 19.81 \\
Phi-4-Reasoning-Vision & 11.54 & 10.39 & 6.71 & 20.16 & 11.12 & 7.83 & 3.93 & 6.07 & 2.08 & 9.54 & 1.03 & 14.24 & 4.90 & 25.13 & 17.80 & 17.27 \\
LLaVA-OneVision 2-8B & 11.55 & 4.00 & 7.15 & 12.14 & 6.42 & 7.22 & 12.92 & 13.23 & 0.00 & 6.41 & 0.19 & 8.62 & 0.48 & 17.07 & 11.34 & 20.71 \\
InternVL3.5-4B & 9.65 & 7.23 & 4.49 & 13.24 & 11.02 & 5.92 & 15.24 & 12.94 & 4.69 & 12.36 & 12.12 & 10.01 & 2.33 & 2.81 & 10.61 & 8.72 \\
Cosmos3-Edge & 9.39 & 7.47 & 8.38 & 17.37 & 4.87 & 7.36 & 3.24 & 2.32 & 0.00 & 8.01 & 1.98 & 11.21 & 14.79 & 24.48 & 14.09 & 22.35 \\
Granite Vision 4.1-4B & 9.23 & 4.44 & 7.84 & 11.97 & 13.92 & 6.04 & 10.09 & 5.64 & 1.56 & 9.12 & 12.45 & 9.91 & 2.41 & 9.14 & 6.78 & 21.60 \\
Mistral Small 4-119B & 6.46 & 4.91 & 2.51 & 5.02 & 6.44 & 2.61 & 11.19 & 1.89 & 0.00 & 3.99 & 2.63 & 3.94 & 6.64 & 5.74 & 1.50 & 0.62 \\
Gemma 4-E2B & 5.22 & 3.81 & 1.52 & 4.03 & 1.82 & 6.04 & 3.08 & 0.46 & 0.00 & 3.82 & 0.01 & 9.24 & 1.54 & 6.78 & 5.06 & 10.03 \\
Ovis2.6-30B-A3B & 3.80 & 4.11 & 5.76 & 3.01 & 3.04 & 1.72 & 6.95 & 2.64 & 13.02 & 5.32 & 1.34 & 1.73 & 0.67 & 3.77 & 1.74 & 0.69 \\
Gemma 4-E4B & 3.04 & 1.45 & 3.49 & 0.53 & 4.26 & 1.48 & 1.41 & 1.81 & 0.00 & 2.40 & 0.28 & 6.58 & 0.74 & 8.22 & 1.25 & 11.97 \\
GPT-6 Astra & 21.83 & 15.85 & 15.60 & 25.33 & 15.61 & 20.18 & 23.09 & 16.35 & 0.00 & 22.66 & 8.94 & 21.51 & 6.51 & 25.77 & 20.42 & 24.44 \\
\bottomrule
\end{tabular}%
}
\end{table}

\begin{table}[!ht]
\centering
\caption{\textbf{Performance across appearance multiplicity.} Each entry is a group-macro percentage within the indicated maximum number of appearances per entity. The final bin produces the clearest rise in Entity NAE and the broadest decline in Score.}
\label{tab:multiplicity_scores_full}
\footnotesize
\begin{tabular*}{\linewidth}{@{\extracolsep{\fill}}lrrrrrrrr@{}}
\toprule
& \multicolumn{4}{c}{Entity NAE$\downarrow$} & \multicolumn{4}{c}{Score$\uparrow$} \\
\cmidrule(lr){2-5}\cmidrule(lr){6-9}
Model & 1 & 2 & 3 & $\geq4$ & 1 & 2 & 3 & $\geq4$ \\
\midrule
Gemini 3.7 Flash & 27.55&26.58&26.29&32.54&40.54&35.27&31.19&30.07\\
Muse-Glimmer-30B & 35.32&31.36&27.56&26.38&28.72&29.68&29.51&33.27\\
Qwen3.8-27B & 35.75&30.19&27.70&36.07&29.98&28.19&30.26&26.61\\
Qwen3.5-397B & 33.20&32.50&25.23&45.92&27.71&27.89&31.36&26.48\\
Gemma 4-31B & 33.47&37.28&34.66&51.63&29.73&24.68&25.27&22.34\\
Qwen3.5-27B & 32.55&35.09&32.26&42.57&27.94&25.44&23.55&27.06\\
Qwen3.5-122B & 32.87&34.29&31.68&44.46&27.75&24.82&26.23&25.93\\
Qwen3.5-35B & 30.84&34.41&28.79&43.00&29.25&23.37&25.07&25.32\\
Gemma 4-26B-A4B & 34.60&39.04&40.33&60.37&28.71&21.84&20.00&17.39\\
Qwen3.5-9B & 35.80&34.98&34.72&45.87&26.53&21.29&24.43&23.82\\
GPT-5.6 Sol & 35.79&38.04&31.50&35.12&21.81&20.55&26.11&25.27\\
InternVL3.5-241B & 41.64&32.31&32.05&41.18&21.72&22.29&17.22&13.12\\
DeepSeek-V4 & 40.22&34.38&32.17&32.38&19.25&21.72&24.16&18.21\\
GPT-5.6 Terra & 37.60&34.95&35.44&41.17&19.96&20.57&19.65&22.23\\
Gemma 4-12B & 35.49&45.32&55.32&71.06&23.19&18.42&11.02&11.04\\
Qwen3.5-4B & 36.90&38.48&40.61&57.80&21.03&16.88&19.88&17.24\\
InternVL3.5-30B & 42.41&36.40&31.11&46.79&18.12&18.63&15.15&12.73\\
MiniMax-M3 & 40.60&37.80&34.21&44.54&17.22&17.45&22.12&14.17\\
GPT-5.6 Luna & 38.82&41.64&45.85&58.41&15.13&17.39&22.60&16.42\\
InternVL3.5-38B & 40.58&38.85&37.12&44.19&15.90&15.25&16.02&15.48\\
InternVL3.5-14B & 39.21&39.05&40.11&49.94&16.58&14.37&16.34&13.32\\
Molmo2-4B & 43.20&36.65&30.22&35.96&13.28&14.20&22.50&21.11\\
Ovis2.6-80B & 45.67&35.02&36.34&44.83&13.82&15.61&17.15&13.55\\
MiniCPM-V 4.5 & 44.47&34.79&30.45&46.91&13.43&15.67&18.13&13.67\\
Cosmos3-Super & 39.58&35.33&33.04&44.82&16.36&13.82&17.60&10.24\\
InternVL3.5-8B & 39.68&39.93&43.33&60.74&18.29&12.45&12.02&9.93\\
Molmo2-8B & 41.66&35.27&30.21&39.70&12.80&11.85&20.31&16.72\\
ERNIE 4.5 VL & 39.98&43.10&39.32&62.92&15.54&12.11&12.41&8.71\\
MiniCPM-V 4.6 & 45.95&40.04&43.45&55.86&14.84&11.64&9.98&8.37\\
Cosmos3-Nano & 43.71&38.00&38.79&53.31&15.74&9.18&14.85&11.87\\
Phi-4 RV & 45.59&46.36&46.67&60.54&12.99&10.26&11.58&9.52\\
LLaVA-OV2-8B & 38.17&42.89&44.14&58.90&12.66&8.58&12.80&11.30\\
InternVL3.5-4B & 47.72&47.29&49.68&63.10&12.43&7.41&11.00&8.87\\
Cosmos3-Edge & 44.14&41.70&43.70&53.39&12.34&7.20&11.32&6.94\\
Granite Vision-4B & 45.33&41.91&48.83&65.31&10.70&7.76&8.61&9.97\\
Mistral Small 4-119B & 50.91&49.36&61.22&82.80&6.36&6.58&3.50&1.16\\
Gemma 4-E2B & 51.67&54.13&59.78&78.89&6.14&4.39&4.64&1.03\\
Ovis2.6-30B & 67.45&68.51&66.22&53.05&2.74&3.94&7.62&4.38\\
Gemma 4-E4B & 59.68&63.96&71.50&81.94&4.02&2.62&1.57&1.16\\
GPT-6 Astra & 33.39&38.28&42.15&49.65&21.97&19.67&22.14&21.27\\
\bottomrule
\end{tabular*}
\end{table}

\begin{table}[!ht]
\centering
\caption{\textbf{Identity correspondence within and across image-formation modes.} Both conditions are reweighted to the same appearance-multiplicity distribution. Positive $\Delta$ values indicate higher error or Score in the cross-mode condition.}
\label{tab:cross_mode_scores_full}
\footnotesize
\begin{tabular*}{\linewidth}{@{\extracolsep{\fill}}lrrrrrr@{}}
\toprule
& \multicolumn{3}{c}{Correspondence NAE$\downarrow$} & \multicolumn{3}{c}{Score$\uparrow$}\\
\cmidrule(lr){2-4}\cmidrule(lr){5-7}
Model & Within & Cross & $\Delta$ & Within & Cross & $\Delta$\\
\midrule
Gemini 3.7 Flash&30.08&35.65&+5.57&29.73&34.51&+4.78\\
Muse-Glimmer-30B&32.81&42.67&+9.86&23.87&30.54&+6.67\\
Qwen3.8-27B&24.17&45.03&+20.86&27.14&28.54&+1.40\\
Qwen3.5-397B&39.34&45.02&+5.68&24.03&28.49&+4.46\\
Gemma 4-31B&41.23&42.66&+1.43&16.34&25.06&+8.72\\
Qwen3.5-27B&25.29&48.37&+23.08&24.88&25.66&+0.78\\
Qwen3.5-122B&43.87&48.80&+4.92&24.00&25.40&+1.39\\
Qwen3.5-35B&38.86&51.29&+12.43&17.86&24.29&+6.43\\
Gemma 4-26B-A4B&48.94&46.93&$-2.01$&13.30&21.55&+8.25\\
Qwen3.5-9B&42.81&53.54&+10.73&22.88&22.18&$-0.70$\\
GPT-5.6 Sol&34.72&50.73&+16.00&22.16&21.86&$-0.29$\\
InternVL3.5-241B&49.46&45.81&$-3.65$&9.58&21.22&+11.64\\
DeepSeek-V4&40.29&45.96&+5.67&12.24&22.10&+9.86\\
GPT-5.6 Terra&36.27&50.22&+13.95&18.23&20.95&+2.73\\
Gemma 4-12B&47.22&48.19&+0.97&9.42&17.15&+7.73\\
Qwen3.5-4B&29.29&57.06&+27.78&13.42&17.56&+4.14\\
InternVL3.5-30B&50.01&50.08&+0.06&8.60&18.04&+9.44\\
MiniMax-M3&46.66&50.55&+3.89&10.44&17.93&+7.49\\
GPT-5.6 Luna&31.79&53.71&+21.93&19.81&17.94&$-1.87$\\
InternVL3.5-38B&55.73&54.91&$-0.82$&8.39&15.77&+7.38\\
InternVL3.5-14B&54.41&53.93&$-0.49$&12.03&14.69&+2.66\\
Molmo2-4B&48.83&62.08&+13.25&10.42&16.25&+5.82\\
Ovis2.6-80B&57.05&56.83&$-0.23$&4.91&16.07&+11.16\\
MiniCPM-V 4.5&45.79&55.81&+10.03&7.70&16.15&+8.44\\
Cosmos3-Super&55.86&63.83&+7.97&9.89&14.07&+4.18\\
InternVL3.5-8B&58.65&57.90&$-0.75$&7.28&12.32&+5.04\\
Molmo2-8B&52.57&67.10&+14.53&10.02&13.58&+3.56\\
ERNIE 4.5 VL&47.78&58.75&+10.97&14.05&11.74&$-2.31$\\
MiniCPM-V 4.6&47.23&56.27&+9.04&11.44&11.15&$-0.29$\\
Cosmos3-Nano&64.89&68.99&+4.10&8.38&10.27&+1.89\\
Phi-4 RV&57.23&61.23&+4.00&8.58&10.46&+1.89\\
LLaVA-OV2-8B&47.27&68.02&+20.76&14.56&9.27&$-5.29$\\
InternVL3.5-4B&57.90&64.63&+6.73&6.10&8.10&+2.00\\
Cosmos3-Edge&65.15&66.78&+1.63&3.19&7.85&+4.66\\
Granite Vision-4B&53.60&67.84&+14.24&6.08&8.26&+2.18\\
Mistral Small 4-119B&67.90&67.83&$-0.07$&3.00&5.69&+2.69\\
Gemma 4-E2B&77.49&65.00&$-12.48$&0.57&4.16&+3.60\\
Ovis2.6-30B&53.26&59.32&+6.06&2.15&4.54&+2.39\\
Gemma 4-E4B&79.17&67.43&$-11.74$&0.93&2.40&+1.47\\
GPT-6 Astra&30.07&46.82&+16.74&24.67&20.11&$-4.55$\\
\bottomrule
\end{tabular*}
\end{table}

\clearpage
\section{Error Paths in Model-Generated Reasoning}
\label{app:reasoning_traces}

A wrong integer does not identify the operation that failed. We therefore examine three cases whose generated reasoning exhibits a different error path: constructing the appearance inventory, linking appearances into physical identities, and applying the requested set operation. For each case, we reproduce the complete reasoning of two representative models in its original order, followed by an analysis of the first consequential divergence. These cases illustrate observable reasoning paths rather than estimate their prevalence, and generated reasoning is not treated as a direct readout of a model's internal computation.

\subsection{A Reflection Need Not Have a Visible Source}

\paragraph{Question.} \emph{Counting both the potted plants physically present on the shelving unit on the left and the reflections of potted plants in the round mirror on the right, how many potted plant appearances are there in the image in total?}
The image contains two direct appearances and three reflected appearances, giving five appearances in total. One reflected plant has no direct counterpart inside the crop.

\begin{figure}[!htbp]
\captionsetup{font=footnotesize,skip=3pt}
\centering
\includegraphics[width=0.96\linewidth]{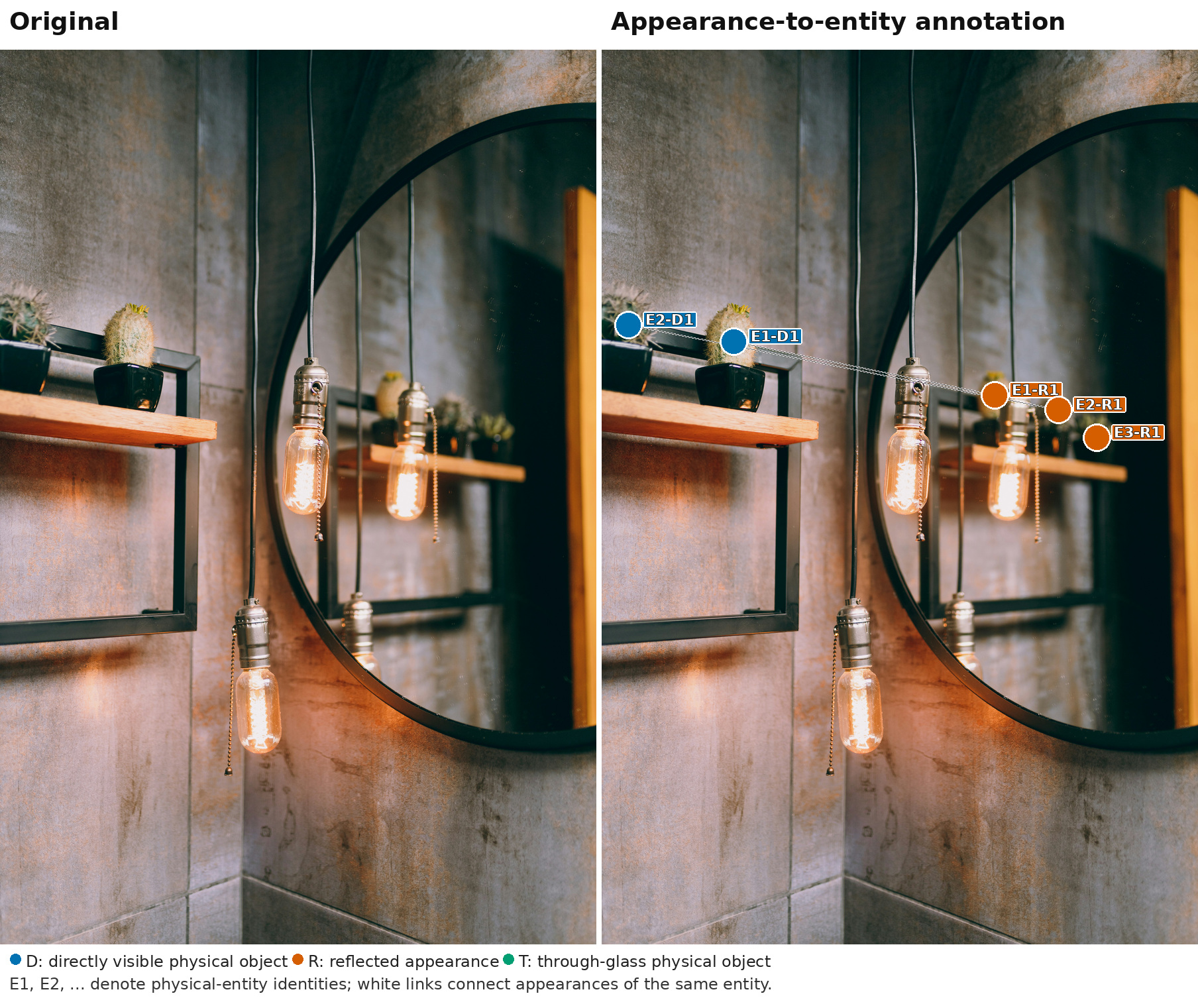}
\caption{\textbf{An appearance inventory that is not pairwise balanced.} The crop contains two directly visible plants but three plant reflections. The unmatched reflection remains an observed appearance even though its physical source is outside the direct view.}
\label{fig:reasoning_case_a}
\end{figure}

\traceentryheader{google/gemma-4-31B-it}{5}{tracecorrect}{CORRECT}
\begin{lstlisting}[style=physcounttrace,title={\tracelistingtitle{MODEL REASONING}}]
Okay, let's count the potted plant appearances in the image.

1.  First, I'll look at the shelving unit on the left. There's one potted plant on the far left: `{"point": [335, 55], "label": "potted plant physically present"}`. 2.  And another one next to it on the same shelf: `{"point": [348, 208], "label": "potted plant physically present"}`. 3.  Now, let's look at the reflections in the mirror. I see a reflection of a potted plant here: `{"point": [397, 660], "label": "reflection of a potted plant"}`. 4.  And another one next to it in the mirror: `{"point": [417, 758], "label": "reflection of a potted plant"}`. 5.  And one more further to the right in the mirror: `{"point": [434, 834], "label": "reflection of a potted plant"}`.

So, there are 2 physically present and 3 reflections. That's a total of 5.
\end{lstlisting}
\begin{lstlisting}[style=physcountresponse,title={\tracelistingtitle{FINAL ANSWER}}]
5
\end{lstlisting}

\traceentryheader{openbmb/MiniCPM-V-4.6-Thinking}{4}{traceincorrect}{INCORRECT}
\begin{lstlisting}[style=physcounttrace,title={\tracelistingtitle{MODEL REASONING AND FINAL ANSWER}}]
Got it, let's count. On the left shelving unit, there are two potted plants: one cactus and one black pot with a plant? Wait, no, looking at the left shelf: there's a black pot with a cactus, and another black pot next to it? Wait, the left shelf has two pots: one with a cactus (potted plant) and one black pot (empty? Or does it have a plant? The image shows two pots on the left shelf: one with a cactus (that's one), and the other black pot is empty? Wait, no, the first pot (leftmost) has a cactus, then the black pot next to it has a small black pot with a plant? Wait, maybe there are two potted plants on the left shelf: the cactus in the first pot and the small plant in the second black pot? Let me check. The left shelf has two pots: one with a cactus (green, spiky), and another black pot in front with a small plant? Or is that just a pot? Hmm, maybe only one potted plant is visibly on the left shelf (the cactus), and then in the mirror, there are reflections of potted plants. The mirror shows reflections of the hanging lights and maybe the shelves, but the potted plants reflected: the shelf on the right has potted plants, including maybe one small plant reflected. Let's list:

Physical potted plants on left shelf: the cactus (1) and maybe the small plant in the black pot? If that small black pot has a small plant, that's 2.

Reflections in mirror: the mirror shows reflections of the hanging lights and possibly the potted plants on the shelf. Looking at the mirror, there are reflections of the potted plants on the shelf (the ones on the left shelf are reflected, and on the right shelf there are potted plants reflected). But the question says "potted plants physically present on the shelving unit on the left" plus "reflections of potted plants in the round mirror".

So physical on left shelf: let's count the actual plants in the left shelf. Left shelf has two pots: one with a cactus (plant), and another pot (black) which seems to have a small plant? Maybe that's a second plant. So that's 2 physical plants.

Then reflections in mirror: the mirror reflects the left shelf, so the potted plants there (the cactus and maybe others) are reflected. But the cactus is physical, its reflection? Or the reflected image includes the plant. But the question says "reflections of potted plants in the round mirror" – so how many are visible as reflected plants? Looking at the mirror, we can see reflections of the hanging lights and also reflections of the shelf with plants, including the small plant in the black pot on the left shelf? Maybe there are at least two reflected plants: the cactus (physical, counted once) and its reflection, plus another plant? This is getting complicated.

Alternatively, count each potted plant that is either physically present (on the left shelf) or its reflection (in the mirror).

Left shelf: physical potted plants: the cactus (1) and the small plant in the black pot (2) – so 2.

Reflections in mirror: the mirror shows reflections of those two, so that's 2 more? But wait, the cactus is physical, its reflection is in the mirror. Similarly, the small plant in the black pot on left shelf has a reflection in the mirror. So total appearances: physical plants (left shelf: 2) + reflected plants (mirror: 2, which are the reflections of those 2) – but that would be 4? But no, the physical plants on left shelf are 2 (cactus and small plant), their reflections in mirror would be two more, so total 4? But that seems off.

Wait, let's list all visible potted plants:

1. Physical on left shelf: cactus (leftmost pot) – 1 2. Black pot next to cactus: small plant? (second pot) – 1 (total 2) 3. Reflections in mirror: the small plant in the black pot on left shelf is reflected, so that's the same plant? Or two plants? The mirror shows reflections of the shelf, including the small plant (so that's one reflection of that plant), and maybe the cactus reflected? The mirror reflects the entire scene, so the cactus's reflection would be visible, and the small plant's reflection.

But the question is: "how many potted plant appearances are there in the image in total?" – that means count every instance of a potted plant, whether physically present or reflected.

So physically present: left shelf has 2 potted plants (cactus and the small plant in black pot) – 2.

Reflected in mirror: those same 2 potted plants are reflected, so that's 2 more (the reflections count as appearances? Or do we count reflections as separate appearances? If a physical plant appears in the mirror, that's a reflection, which is still an appearance of that plant. So each physical plant has one appearance (on shelf) and one appearance (reflected in mirror). So if there are 2 physical plants, each has 1 appearance on shelf and 1 in mirror, total 4? But that doesn't make sense.

Alternatively, count the actual plants: left shelf has 2 potted plants (cactus and small plant) – 2.

Mirror reflects: those 2 plants are reflected, so in the mirror, we see those 2 plants (reflected), which are the same as the physical ones, so total appearances: physical ones (2) + reflected ones (2) = 4? But that would be 4.

Wait, let's look at the image: left shelf has two pots: one with cactus, one with small plant (black pot). The mirror has a round frame, and reflections show those two pots with plants (so the plants are reflected, meaning they are present in the reflection, so they are "appearing" in the mirror as plants). So the physical plants are 2, and their reflections are 2, so total appearances (physical + reflected) is 4? But that would be 2 physical + 2 reflected = 4.

But wait, the cactus is one physical plant, its reflection is a second appearance. The small plant in the black pot is another physical plant, its reflection is a third appearance. So total appearances: 4 (physical: 2, reflected: 2).

But physical plants on left shelf: 2 (cactus and small plant) – physical appearances: 2.

Reflections in mirror: the mirror shows 2 reflections (one for each physical plant), so reflected appearances: 2.

Total: 2 + 2 = 4.

Yes, that makes sense. So total appearances (physical present + reflected in mirror) are 4. </think>

4
\end{lstlisting}

\paragraph{Analysis.}
Gemma's generated reasoning builds the answer from observations: it enumerates two direct plants, independently enumerates three reflections, and only then adds the two sets. MiniCPM's rationale reverses that dependency. After finding two direct plants, it assumes that the mirror must contain one copy of each and fixes the reflected count at two. The third reflected plant disappears from the final tally because it violates this assumed one-to-one mapping.

The recorded mistake comes from treating a reflection as a duplicate of the visible crop rather than as another view of the surrounding scene. A mirror may reveal a source outside the camera frustum, omit a visible source because of its orientation, or show one source in several panels. None of these cases preserves the direct count. In this example, the reflected regions must remain in the appearance inventory before their physical sources are considered; the MiniCPM rationale instead uses a one-to-one identity assumption to revise the tally.

\subsection{Entity Counting Depends on the Identity Edges}

\paragraph{Question.} \emph{How many different shoes are actually in the scene captured in this photo?}
Four shoe appearances form two direct--reflection pairs. The answer is two physical shoes.

\begin{figure}[!htbp]
\captionsetup{font=footnotesize,skip=3pt}
\centering
\includegraphics[width=0.96\linewidth]{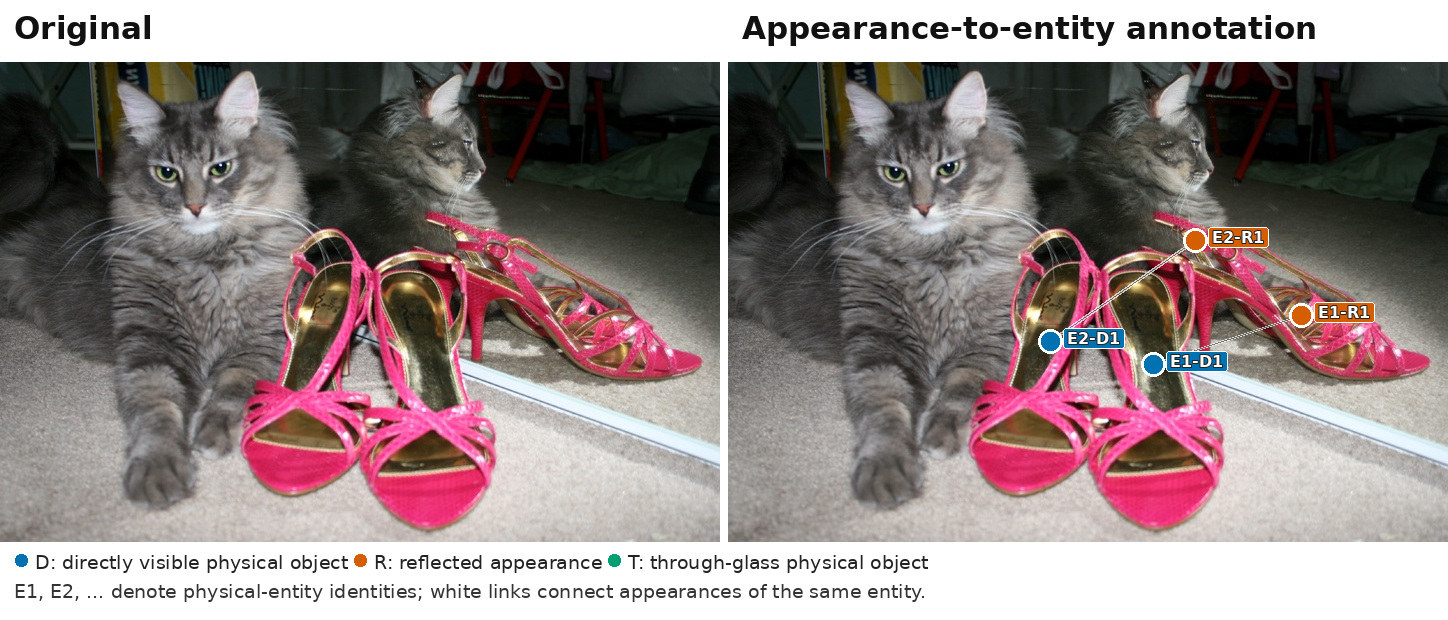}
\caption{\textbf{Four shoe appearances connected into two identities.} The mirror boundary does not remove visual evidence; it determines which observations should be linked before the entities are counted.}
\label{fig:reasoning_case_e}
\end{figure}

\traceentryheader{google/gemma-4-31B-it}{3}{traceincorrect}{INCORRECT}
\begin{lstlisting}[style=physcounttrace,title={\tracelistingtitle{MODEL REASONING}}]
Okay, let's count the shoes in the image.

1.  First, there's a pink high-heeled sandal on the left: `{"point": [684, 435], "label": "pink high-heeled sandal"}`. 2.  Next to it is another pink high-heeled sandal in the middle: `{"point": [727, 621], "label": "pink high-heeled sandal"}`. 3.  And then there's a third pink high-heeled sandal lying on its side to the right: `{"point": [551, 816], "label": "pink high-heeled sandal"}`.

So, I count 3 shoes in total.
\end{lstlisting}
\begin{lstlisting}[style=physcountresponse,title={\tracelistingtitle{FINAL ANSWER}}]
3
\end{lstlisting}

\traceentryheader{OpenGVLab/InternVL3\_5-241B-A28B}{2}{tracecorrect}{CORRECT}
\begin{lstlisting}[style=physcounttrace,title={\tracelistingtitle{MODEL REASONING}}]
Okay, let's see. The image shows a fluffy gray cat lying on the floor next to a pair of bright pink high-heeled shoes. The shoes are placed in front of a mirror, which is reflecting the scene. The cat's reflection is also visible in the mirror.

The question is asking how many different shoes are actually in the scene. So, the key here is to determine how many physical shoes are present, not counting the reflections. The mirror shows a reflection of the shoes, which might make it look like there are two pairs, but in reality, there's only one pair in front of the mirror.

Looking at the image, the shoes in front of the mirror are a pair, so that's two shoes. The mirror reflects them, making it appear as if there's another pair behind, but that's just a reflection. So the actual number of different shoes (i.e., individual shoes) present in the scene is two. However, the question might be phrased to ask for the number of pairs, but the wording says "different shoes," which could mean individual shoes. Wait, the wording says "how many different shoes are actually in the scene." So each shoe is a separate one. But since they are a pair, they are two individual shoes. But maybe the answer is one pair, but the question says "different shoes," so maybe two. Alternatively, maybe the reflection is considered not part of the actual scene. So the actual shoes are two, one left and one right. The mirror shows their reflection, but those aren't real. So the answer is two. Wait, but the image shows the shoes in front of the mirror. The mirror reflects them, but the actual shoes are two. So the answer is two. But maybe the user is tricking us into thinking that the reflection is a different shoe. But no, the actual shoes are two. So the integer answer is 2. Wait, but looking at the image again, the shoes are placed in front of the mirror. So the mirror shows a reflection of the shoes, but the actual shoes are two. So total different shoes (individual) is two. So the answer is 2. Alternatively, maybe the question is a trick, but I think the answer is two. Wait, but the image shows two shoes (a pair) and their reflection. So actual shoes are two. So the answer is 2. But maybe the question is phrased as "different" in terms of pairs. Hmm. Alternatively, maybe the shoes are the same, so "different" might refer to unique shoes. But they are a pair, so same design. But the question is about how many different shoes are present. If "different" refers to individual shoes, then two. If it's asking how many pairs, then one. But the question says "different shoes" and "exactly one integer." Let me check the image again. The shoes are a pair of pink high heels. So two shoes. The mirror reflects them, but that's not real. So the answer is 2. So the integer is 2. But wait, maybe the user is asking how many unique shoes, like left and right. But they are a pair, so two. So answer 2. Alternatively, maybe the answer is 1, considering a pair as one item. But the question says "shoes" plural. Hmm. Let me think again. The question says "how many different shoes are actually in the scene captured in this photo?" So each shoe is a separate entity. So two shoes. The mirror image is not a real shoe. So answer 2. So the integer is 2. But maybe I'm misunderstanding. Let me check the image again. The shoes are placed in front of a mirror, so the mirror shows their reflection. So in the photo, you can see two shoes and their reflection. But the actual physical shoes are two. So the answer is two. So the integer is 2. Alternatively, if the question is a trick, but I think that's correct.
\end{lstlisting}
\begin{lstlisting}[style=physcountresponse,title={\tracelistingtitle{FINAL ANSWER}}]
2
\end{lstlisting}

\paragraph{Analysis.}
Gemma's rationale stops after detecting three salient shoe regions. InternVL instead explains why the regions are repeated: the shoes lie against a mirror, and the rear-looking pair is generated by reflection. The observable difference is not simply more careful arithmetic. InternVL changes the unit being counted from image regions to connected identity groups.

This distinction is easy to lose because visual similarity alone does not determine identity. Matching color and design are compatible with both a reflected shoe and another shoe from the same product line. The relevant evidence is relational: the mirror boundary, reversed placement, and the alignment between direct and reflected regions. InternVL's rationale uses these cues to introduce two cross-boundary identity links, whereas Gemma's does not. Expressed in the benchmark's annotation graph, their stated interpretations disagree on the edges and therefore on the number of connected components.

\subsection{Physical-Only Is a Set Difference, Not a Physical Count}

\paragraph{Question.} \emph{How many people appear only in the physical space of the stage, with no reflection in the mirror?}
The directly visible dancer also has reflected appearances. The physical-only set is empty, so the answer is zero.

\begin{figure}[!htbp]
\captionsetup{font=footnotesize,skip=3pt}
\centering
\includegraphics[width=0.96\linewidth]{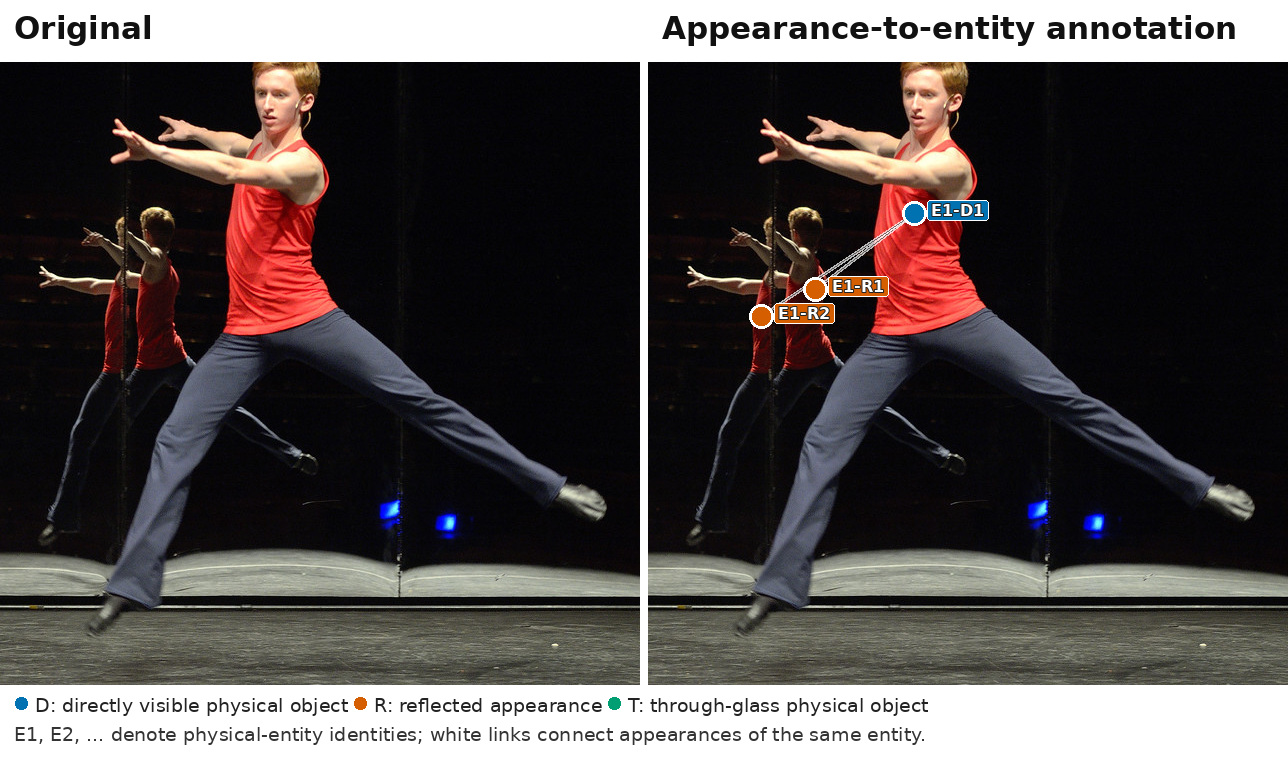}
\caption{\textbf{A nonempty physical set but an empty physical-only set.} The direct dancer and the two mirror appearances belong to one identity; the exclusion condition removes that identity from the answer.}
\label{fig:reasoning_case_cset}
\end{figure}

\traceentryheader{google/gemma-4-31B-it}{0}{tracecorrect}{CORRECT}
\begin{lstlisting}[style=physcounttrace,title={\tracelistingtitle{MODEL REASONING}}]
Okay, let's look at the image to find how many people are in the physical space of the stage but *not* reflected in the mirror.

1.  First, I identify the people. I see one main person in a red tank top and dark blue pants, leaping across the stage. 2.  Then, I see reflections in the mirror behind him. There are two reflections of a person in a red tank top and dark blue pants. 3.  Wait, looking closer at the reflections: there's one reflection that corresponds directly to the main person's position and pose (the one closest to him). 4.  There's another reflection further back, which seems to be of another person who might be off-camera or further back on the stage. 5.  Wait, let's re-examine. There's one person physically on the stage (the main subject). 6.  There are two reflections. One is a clear reflection of the person jumping. The other reflection is further to the left, and it doesn't seem to have a corresponding physical person visible in the frame. 7.  However, the question asks for "people [who] appear *only* in the physical space of the stage, with *no* reflection in the mirror". 8.  The main person on stage *is* reflected in the mirror. 9.  Are there any other people on the stage who are *not* reflected? No, I only see one physical person. 10. Let's double check. The person leaping is clearly reflected. There's another figure in the reflection, but there's no physical person corresponding to them in the visible part of the stage. 11. So, the only person on the stage has a reflection. Therefore, there are zero people who appear only on the stage with no reflection.

Let me re-read: "How many people appear only in the physical space of the stage, with no reflection in the mirror?" - Person A (leaping): Appears on stage AND has a reflection. - Person B (reflected): Appears in the reflection, but not on the physical stage in the image.

So, none of the people visible on the stage are without a reflection. Wait, let me look again. Maybe there's someone else? Looking at the whole image... I see one main person. I see two reflections. One reflection is definitely him. The other reflection is another person in a red top. Where is that person physically? They are not visible in the non-mirrored part of the image. So, the main person is reflected. There's no one else on the stage who is not reflected.

Therefore, the number is 0.
\end{lstlisting}
\begin{lstlisting}[style=physcountresponse,title={\tracelistingtitle{FINAL ANSWER}}]
0
\end{lstlisting}

\traceentryheader{Qwen/Qwen3.8-27B}{1}{traceincorrect}{INCORRECT}
\begin{lstlisting}[style=physcounttrace,title={\tracelistingtitle{MODEL REASONING}}]
The user wants a count of people who appear only in the physical stage space, not as mirror reflections.

Looking at the image, I see a stage with a mirror or reflective surface. There's a main dancer in the foreground wearing a red tank top and dark blue pants, leaping with arms extended.

Behind him, in what appears to be a mirror reflection, there are two figures also in red tops and dark pants, mirroring the pose. Wait, let me look carefully.

The image shows a dancer in the foreground (the main subject). Behind him, there's a vertical line that looks like a mirror edge. To the left of that line, there are two reflected figures? Let me look.

Actually, the setup: there's a mirror on the back wall. The dancer in the foreground is the real person. The reflections in the mirror show his reflection. But I see two figures in the background on the left side.

Let me parse the geometry. The foreground dancer's body is oriented with his left leg extended to the right (his right leg?), his torso facing us, arms out.

In the background left, I see one figure with arms out (red top, dark pants) and another figure partially overlapping. Hmm.

Let's look at the reflections. The mirror is likely the dark vertical surface behind. The reflection of the dancer would appear in the mirror. The dancer is leaping to the right (his left). The reflection would be reversed.

In the background, I see two red-topped figures. One is more complete on the left (with arms out to the left), and one is in the middle-left (the one whose body aligns with the foreground dancer's torso). Actually the middle-left figure looks like a reflection of the foreground dancer's upper body, and the leftmost figure is the reflection of the lower body? That doesn't make sense.

Let's think about mirror geometry. Suppose there are two mirrors at an angle, producing multiple reflections. Or maybe there's just one mirror and we see the dancer's reflection plus a second dancer's reflection? But the question asks how many people appear ONLY in physical space, i.e., not as reflections. So we need to count the real people on stage who do not have a corresponding reflection? Or count real people, excluding reflections?

The phrasing: "How many people appear only in the physical space of the stage, with no reflection in the mirror?" This means count the people who are physically present on stage (real people), as opposed to reflections. But "with no reflection in the mirror" might mean we should count people that we see as physical (not as mirror images). Hmm, but every physical person would have a reflection in the mirror. So "appear only in the physical space" means we look at the image and identify which figures are physical and which are reflections. The physical ones are the answer.

So the question reduces to: how many real dancers are on stage (as opposed to their mirror images)?

Looking at the image, the foreground dancer is clearly the real one (he's in focus, lit, in front). The background figures are reflections.

But wait, there are two background figures. If there were only one real dancer, his single reflection in a single flat mirror would produce one image. Why two?

Possibility: There are two mirrors meeting at an angle (like a corner), producing two reflections of the same dancer. Or the background shows two reflections of the same dancer from two different mirror panels.

Look at the vertical line in the middle of the image (around x=620). That's the edge between two mirror panels perhaps. The right side is dark (no reflection visible there because the angle doesn't show the dancer, or it's a black curtain). The left side is the mirror showing the reflection.

Hmm, but I see two figures on the left side of that vertical line. Let's look at their positions. One figure (call it A) is at the far left, arms extended to the left, body leaning, legs together-ish. Another figure (call it B) is just left of center, with one arm extended up-left and torso in red, and a leg extended to the right (the leg at around (430,620)).

Actually, figure B's leg extended to the right (the dark pant leg ending in a shoe at ~(480,630)) might be the reflection of the foreground dancer's extended leg? The foreground dancer's extended leg goes to the right (to ~(900,700)). The reflection of that leg in a vertical mirror behind him would go to the left, not right. Hmm.

This is getting too detailed. Let's think about the intended answer. The question is likely from a VQA dataset testing reflection understanding. The expected answer is probably 1: there is one real dancer on stage; all other figures are reflections.

But the phrase "with no reflection in the mirror" is odd. Maybe it means: count the people such that the person themselves is in physical space (not a reflection). I.e., how many physical people do we see? That's 1.

But maybe there are two physical dancers, and the mirror shows reflections of both, making it confusing. Let's look: foreground dancer is real. Is there a second real dancer hidden? The background figures are dimmer and look like reflections (they're behind the mirror line, less lit). The leftmost figure A and middle figure B could be two reflections of the one real dancer from two mirror surfaces.

Actually, look at the geometry: The real dancer's body: his right arm (our left) is extended forward-left, his left arm (our right) extended forward-right? No, both his arms are extended to our left actually. Look: his shoulders, one arm goes to our left-up (the hand at ~(200,120)), the other arm also to our left but more horizontal (hand at ~(270,110)). Both arms point to our left. His body is twisted. His right leg (our right) is extended to our right. His left leg (our left) is bent, planted? Actually his left leg (our left) is extended down-left to the foot at ~(160,880). So both legs spread: one to our left-down, one to our right-down. It's a leap/split jump.

Now a vertical mirror behind him (the dark backdrop) would reflect him. The reflection's arms would point to the right (mirror flips horizontally). But the background figures have arms pointing to our left as well. Hmm, that suggests they are not simple horizontal flips, or the mirror is angled.

This is too much. The likely intended simple answer: 1.

Let's consider the possibility the answer is 1 because only the main dancer is "physical" and the rest are reflections. The question's wording "appear only in the physical space of the stage, with no reflection in the mirror" might be a clumsy way of saying "count the real people (not the mirror images)." So answer = 1.

But wait, maybe the question is the opposite: count people who appear ONLY in physical space, meaning we should look at the set of "people" depicted and determine which ones are not reflections. The reflections are "appear in the mirror." The physical people "appear in physical space." So count physical = 1.

Alternatively, maybe there are 2 physical people: the main dancer and a partner off to the side that we see only physically, whose reflection isn't visible. But I don't see a second clear physical person. The two background figures are both in the mirror region (left of the vertical seam). They look like reflections.

Let me look at the lighting: foreground dancer is brightly lit. Background figures are also fairly lit but appear behind a glass line. The vertical line at x~190 and x~620 might be the edges of a mirror panel? Actually there's a vertical line at around x=190 (a thin bright line) and at x=620. These could be the frame of a mirror that occupies the middle region (x from 190 to 620). In that middle region we see figure B (the reflection of the foreground dancer). To the left of x=190, we see figure A, which is on a different mirror panel? Or figure A is the reflection on a side mirror?

Hmm, if the mirror occupies x in [190,620], then figure B is the reflection of the foreground dancer in that mirror. Then what is figure A? Figure A at x in [60,280] partially overlaps the mirror region and partially left of it. Figure A's body is to the left.

I think the scene is: a single large mirror on the back wall, but the photo captures the dancer's reflection, and due to the dancer's pose, the reflection appears as one figure. But we see two figures in the background, which is puzzling.

Let's look very carefully at the background. I'll describe pixel regions.

Figure A (leftmost): head at ~(185,260), red top torso at ~(180,380), arms extended to upper-left (hands at ~(140,275) and ~(70,340)), legs in dark pants going down-left to foot at ~(75,750). So figure A is a full-body figure with arms to the left.

Figure B (middle): head at ~(240,255), red top torso at ~(240,400), one arm extended to upper-left (hand at ~(185,300)? or that's figure A's?), and a leg extended to the right (the dark pant leg from ~(350,500) to (480,630)). Figure B's other leg? Not clearly visible, maybe overlapping with foreground dancer's leg.

Hmm, so figure B's extended leg goes to the right (our right). The foreground dancer's extended leg also goes to our right (to 900,700). If the mirror is behind the dancer, the reflection's extended leg would go to our left. So figure B's right-going leg doesn't match a simple reflection.

Wait, maybe figure B is actually the real second dancer? But figure B is dimmer and behind the vertical line? The vertical line at x=190 passes through figure A's torso. The vertical line at x=620 is to the right of figure B.

I'm overanalyzing a low-res image. Let's think about typical VQA groundings. This image likely from a "reflection" reasoning set. The question: "How many people appear only in the physical space of the stage, with no reflection in the mirror?" The expected answer is likely 1.

But let's double check by thinking: maybe the trick is that the foreground dancer's reflection is figure B, and figure A is the reflection of a second real dancer who is mostly out of frame? No.

Alternatively maybe the answer is 0? No, the foreground dancer is physical.

Let's think about "appear only in the physical space ... with no reflection in the mirror." Could be interpreted as: among all the person-figures visible in the image, how many of them are the kind that exist only physically (i.e., are real people) and not also present as a reflection? But a real person also appears as a reflection, so they appear in both. The phrase "with no reflection in the mirror" might modify "appear only in the physical space" meaning we consider the physical-space appearances that don't have a corresponding mirror appearance? That's weird.

Maybe simpler: The image contains person-figures. Some are real (on stage), some are reflections (in mirror). The question asks how many of the figures are real (physical). The reflection figures are "in the mirror." The real figures are "in physical space." So count real = ?

If there's one real dancer and his reflection produces the background figures, then real count = 1.

But why two background figures then? Possibly because there are two mirrors (the stage has a mirrored wall at an angle, like a corner reflector), producing two reflections of the one dancer. That's common in dance sets to create multiple images. So one dancer -> one physical + two reflections = 3 figures total. The question: physical only = 1.

Yes I think that's it. The set designer uses angled mirrors to multiply the dancer's image. So we see 3 figures: 1 real, 2 reflections.

Thus answer = 1.

Let's sanity check the visual: real dancer in front, bright. Two dimmer figures behind, consistent with reflections. Yes.

Thus the number of people appearing only in physical space = 1.

The question wants exactly one integer. So output 1.

I'm fairly confident. Let's also consider if maybe the foreground is the reflection and the background is real? No, foreground is clearly in front, on the stage floor with shadow, bright. Background is in the dark mirror area.

Thus answer 1.

I'll output 1.
\end{lstlisting}
\begin{lstlisting}[style=physcountresponse,title={\tracelistingtitle{FINAL ANSWER}}]
1
\end{lstlisting}

\paragraph{Analysis.}
Qwen's rationale describes the relevant visual structure: one foreground dancer, background reflections, and multiple mirror panels as the source of repetition. The answer changes after the question is paraphrased as the number of ``real dancers on stage.'' The stated computation then becomes $|P|$ although the question asks for $|P\setminus R|$, where $P$ contains identities with direct appearances and $R$ contains identities with reflected appearances.

Gemma's rationale retains the exclusion predicate. It first places the directly visible dancer in $P$, then places the same identity in $R$, and removes it. The answer is zero even though the physical set is nonempty. The two recorded rationales therefore differ at the requested operation: Qwen's scene description is largely sufficient, but its final count is not licensed by that description.

The unsuccessful trace repeatedly strengthens the correct claim that there is one physical dancer, yet never restores the discarded phrase ``with no reflection.'' In this case, additional reasoning increases confidence without improving semantic fidelity. The missing operation, rather than the amount of visual inspection, determines the wrong answer.

\paragraph{Implication for evaluation.}
The cases establish a narrower but important point: the same final counting error can accompany different observable reasoning failures. In the plant rationale, a one-to-one reflection assumption changes the appearance tally; in the shoe rationale, the stated identity links are incomplete; in the dancer rationale, the exclusion operator is dropped after the scene has been described. Aggregate integer metrics identify which question families and scene structures are difficult, but they cannot distinguish these paths on their own.

The joint-presentation result in Section~\ref{sec:detailed_analysis} is compatible with reuse of
appearance and identity information across related questions, but it does not
identify that reuse as the causal internal mechanism. Directly separating
appearance detection, identity linking, and set operations would require
structured predictions for the appearance nodes, formation labels, identity
edges, and requested operation. The present traces serve instead as qualitative
examples of why those intermediate outputs would make future diagnosis more
precise.

\end{document}